\documentclass[lettersize,journal]{IEEEtran}
\usepackage{amsmath,amsfonts,amssymb}
\usepackage{algorithmic}
\usepackage{algorithm}
\usepackage{booktabs}
\usepackage{multicol}
\usepackage{multirow}
\usepackage{booktabs}
\usepackage{enumerate}
\usepackage{xcolor}
\usepackage{colortbl}
\usepackage{tabularray}
\usepackage{microtype}
\usepackage[utf8]{inputenc}
\usepackage[T1]{fontenc}
\UseTblrLibrary{booktabs}
\usepackage{tabularx}
\usepackage{hhline}
\usepackage{bbding}
\usepackage{tikz}
\usepackage{paralist}
\usepackage{array}
\usepackage[caption=false,font=scriptsize,labelfont=sf,textfont=sf]{subfig}
\usepackage{textcomp}
\usepackage{stfloats}
\usepackage{soul}
\usepackage{url}
\usepackage{verbatim}
\usepackage{graphicx}
\usepackage{hhline}
\usepackage{cite}
\usepackage[colorlinks,
linkcolor=green,
anchorcolor=black,
citecolor=green]{hyperref}
\usepackage{orcidlink}
\def\eg{\emph{e.g.}}

\newtheorem{remark}{\bfseries Remark}
\newtheorem{definition}{\bfseries Definition}

\begin{document}

\title{You Only Look Omni Gradient Backpropagation for Moving Infrared Small Target Detection}

\author{Guoyi~Zhang~\orcidlink{0009-0004-2931-6698},~Guangsheng~Xu~\orcidlink{0009-0006-4235-9398},~Siyang~Chen~\orcidlink{0009-0008-9465-6647},~Han~Wang~\orcidlink{0009-0008-2794-2520}~and~Xiaohu~Zhang~\orcidlink{0000-0003-4907-1451}
	\thanks{Manuscript received xxx, xxx; revised xxx, xxx.}
	\thanks{Corresponding authors: \emph{Han Wang and Xiaohu Zhang}}
	\thanks{Guoyi~Zhang, Guangsheng~Xu, Siyang~Chen, Han~Wang and Xiaohu~Zhang are with the School of Aeronautics and Astronautics, Sun Yat-sen University, Shenzhen 518107, Guangdong, China.(email: zhanggy57@mail2.sysu.edu.cn;\\zhangxiaohu@mail.sysu.edu.cn)}
}

\markboth{Journal of \LaTeX\ Class Files,~Vol.~14, No.~8, August~2021}%
{Zhang \MakeLowercase{\textit{et al.}}: You Only Look Omni Gradient Backpropagation for Moving Infrared Small Target Detection}


\maketitle

\begin{abstract}
Moving infrared small target detection is a key component of infrared search and tracking systems, yet it remains extremely challenging due to low signal-to-clutter ratios, severe target-background imbalance, and weak discriminative features. Existing deep learning methods primarily focus on spatio-temporal feature aggregation, but their gains are limited, revealing that the fundamental bottleneck lies in ambiguous per-frame feature representations rather than spatio-temporal modeling itself. Motivated by this insight, we propose BP-FPN, a backpropagation-driven feature pyramid architecture that fundamentally rethinks feature learning for small target. BP-FPN introduces Gradient-Isolated Low-Level Shortcut (GILS) to efficiently incorporate fine-grained target details without inducing shortcut learning, and Directional Gradient Regularization (DGR) to enforce hierarchical feature consistency during backpropagation. The design is theoretically grounded, introduces negligible computational overhead, and can be seamlessly integrated into existing frameworks. Extensive experiments on multiple public datasets show that BP-FPN consistently establishes new state-of-the-art performance. To the best of our knowledge, it is the first FPN designed for this task entirely from the backpropagation perspective.
\end{abstract}


\section{Introduction}
\begin{figure}[!t]
	\centering
	\includegraphics[width=\linewidth]{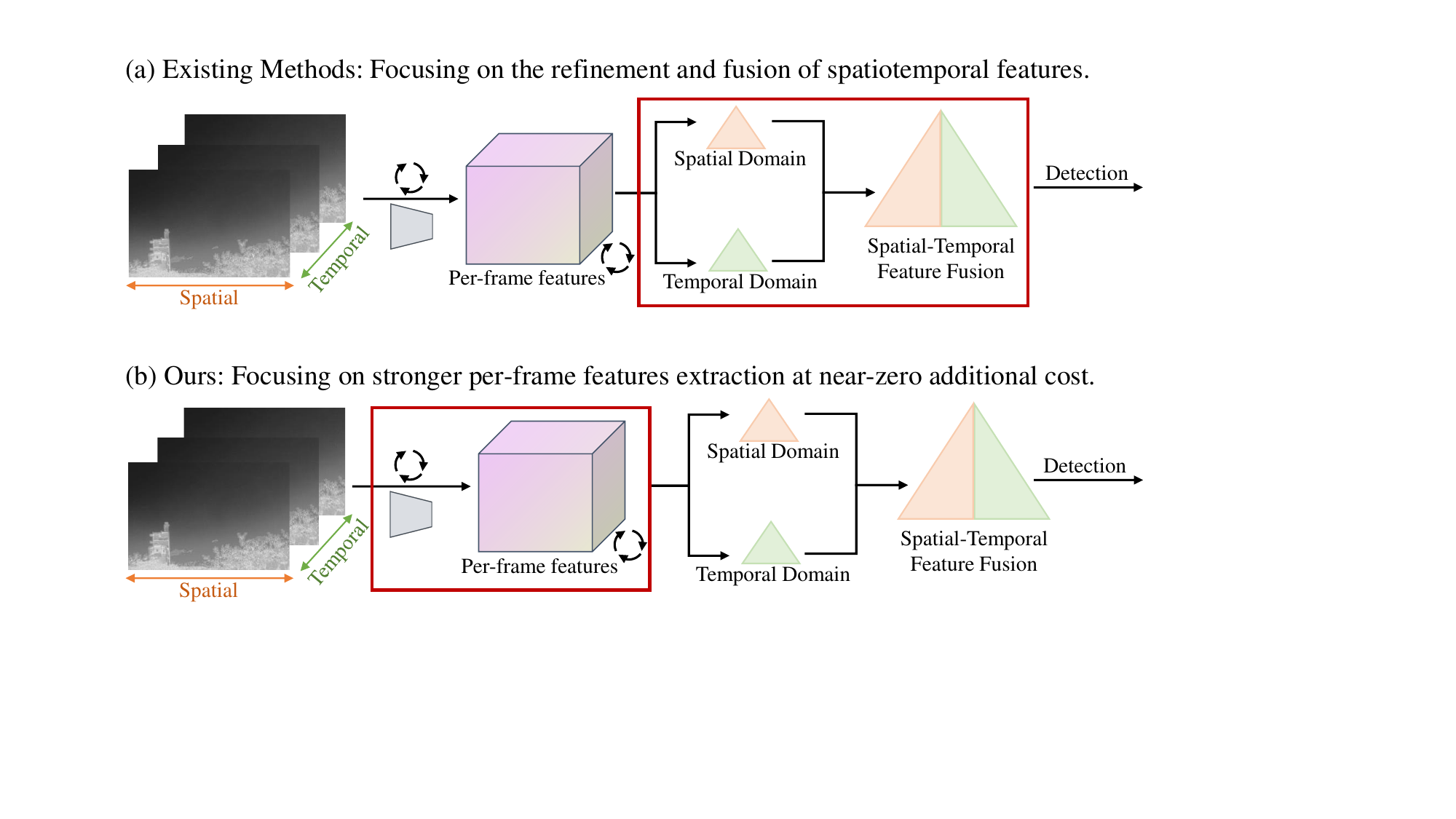}
	\caption{Comparison between our method and existing methods. Unlike existing methods focusing on spatiotemporal feature refinement and fusion, our method is motivated by the insight that the key to video object detection lies in modeling feature-to-feature correlations along the temporal dimension, and thus emphasizes learning stronger per-frame representations. Importantly, from a backpropagation perspective, the proposed method introduces near-zero additional computational overhead and can be seamlessly integrated with existing frameworks, consistently leading to measurable performance gains.}
	\label{fig:Compare}   
\end{figure}

\IEEEPARstart{M}{oving} infrared small target detection serves as the core component of infrared search and tracking (IRST) systems \cite{8734113}, playing a vital role in applications such as missile early warning and military surveillance. However, detecting moving infrared small targets under complex background clutter remains a long-standing challenge due to extreme imbalance between the target and background \cite{11146868}, low signal-to-clutter ratio (SCR) \cite{11156113}, and the absence of distinctive discriminative features \cite{10508299}.
In recent years, deep learning methods have gradually become mainstream owing to their powerful performance and strong generalization ability \cite{MSHNet}. As illustrated in Fig. \ref{fig:Compare}, existing approaches primarily focus on refining and fusing spatio-temporal features \cite{luo2026deformable,chen2024sstnet,duan2024triple}. However, it is \textbf{worth noting} that despite increasingly complex spatio-temporal learning strategies, the performance gains remain marginal \cite{duan2024triple}, which motivates us to rethink the problem of moving infrared small target detection.

Specifically, moving infrared small target detection can be formulated as a video object detection problem \cite{10024313}, where the key challenge lies in modeling temporal features and their inter-feature correlations \cite{10934730}. This observation leads to a \textbf{crucial insight}: if the underlying feature representations are ambiguous, any correlations derived from them will inevitably be unreliable. Moreover, existing methods typically employ Feature Pyramid Networks (FPNs) \cite{lin2017feature} to aggregate multi-level features in a MiSo (multi-input single-output) manner \cite{wang2024yoloh}, producing a single-frame representation. Such a design raises an important question: \textit{can conventional FPNs generate sufficiently robust feature representations for the unique demands of moving infrared small target detection?}

\begin{figure*}[!t]
	\centering
	\includegraphics[width=1\linewidth]{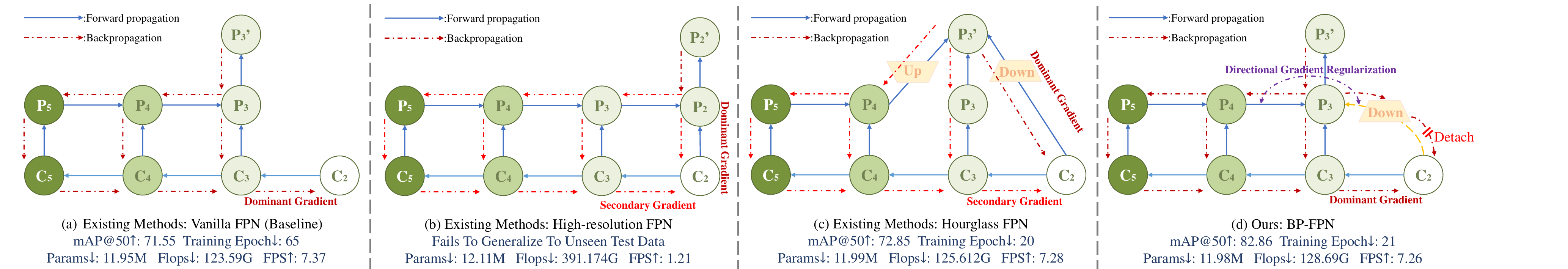}
	\caption{Comparison of the \textbf{Macro Architectures} of Different FPNs. Here, $C_i$ denotes backbone features, $P_i$ represents FPN features, and $P_i^{'}$ is the final output feature. The quantitative results are obtained on the IRDST dataset. (a) The vanilla FPN is commonly employed in current moving infrared small target detection models. It features low computational cost and memory footprint, making it suitable for inter-frame feature aggregation in videos. However, it suffers from information loss for small objects, which not only degrades detection performance but also slows down convergence. (b) The High-Resolution FPN employs shortcut connections that directly transmit low-level semantic cues, which compromises its generalization to unseen environments and significantly increases computational burden while lowering inference speed. (c) The Hourglass FPN simultaneously fuses features of different resolutions to strengthen multi-scale representation and small-object localization, but it remains affected by the optimization difficulties of FPNs. (d) The proposed BP-FPN is designed from the perspective of backpropagation, achieving significant performance improvement with negligible additional complexity.}
	\label{fig:Compared_With_Ex}
\end{figure*}

As illustrated in Fig.~\ref{fig:Compared_With_Ex}, moving infrared small target detection requires spatio-temporal information aggregation. Most existing methods adopt the feature aggregation strategy in subfigure (a), which overlooks crucial low-level semantic information in high-resolution feature maps, resulting in the loss of important target details. In contrast, directly incorporating a high-resolution branch \cite{9947071}, as shown in subfigure (b), often leads to significant degradation in generalization performance on unseen test datasets. Gradient flow analysis further reveals that this strategy introduces a shortcut connection between the training outputs and $C_2$ features, encouraging shortcut learning \cite{10250856}. The fusion method \cite{wang2024yoloh} in subfigure (c) faces optimization difficulties and offers limited improvements; however, collaborative feature fusion helps suppress shortcut learning more effectively than in (b).

\begin{figure*}[!t]
	\centering
	\includegraphics[width=1\linewidth]{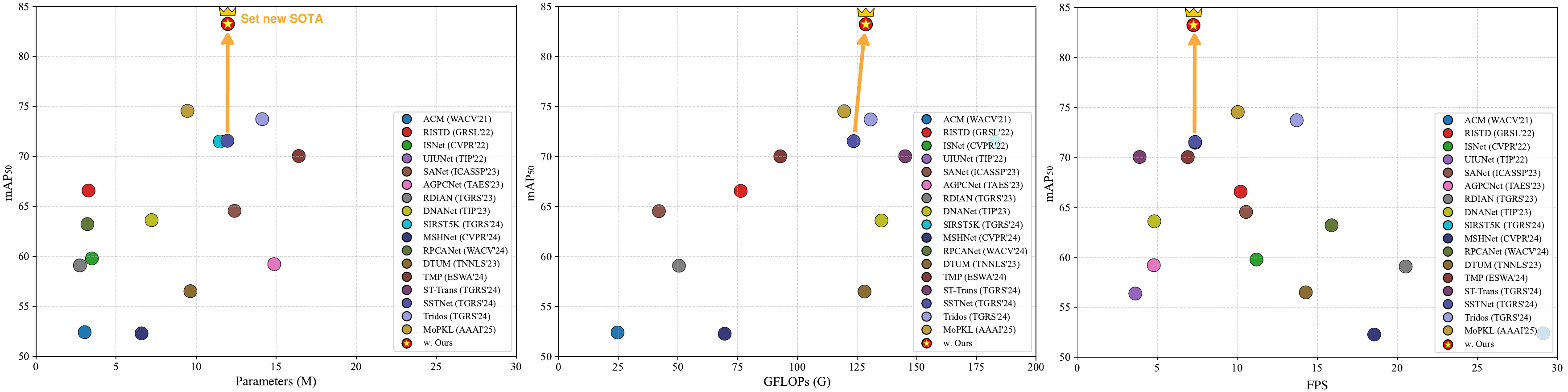}
	\caption{Parameter–$\text{mAP}_{50}$, FLOPs–$\text{mAP}_{50}$ and FPS–$\text{mAP}_{50}$ plots on the IRDST dataset. By integrating our BP-FPN, we achieve a significant performance improvement with near-zero additional computational overhead, establishing a new state-of-the-art (SOTA).}
	\label{fig:Compared_With_mAP}
\end{figure*}

\begin{table}[!t]
	\centering
	\caption{Usability Study of the Plug-and-Play Components (BP-FPN) on the IRDST Dataset. ``R'', ``T'', and ``C'' denote RNN-based spatio-temporal aggregation, Transformer-based spatio-temporal aggregation, and 3D CNN-based spatio-temporal aggregation, respectively.}
	\label{Tab:General}
	\setlength{\tabcolsep}{4.8pt}
	\resizebox{0.48\textwidth}{!}{\begin{tabular}{c|c|c|cc|cc|c}
		\noalign{\hrule height 1pt}
		\textbf{Methods}                    & \textbf{Frames}    & \textbf{Type}             & $\textbf{mAP}_\textbf{50}$$\uparrow$                  & \textbf{F1}$\uparrow$                 & \textbf{Flops}$\downarrow$                 &\textbf{Params}$\downarrow$ & \textbf{FPS}$\uparrow$                   \\ \noalign{\hrule height 1pt}
		SSTNet & 5 & R & 71.55 & 85.11 & 123.59G & 11.95M & 7.37 \\
		w. Ours  & 5 & R & 82.86 & 91.48 & 128.69G &  11.98M & 7.26 \\ \noalign{\hrule height 1pt}
		Tridos & 5 & T & 73.72  & 86.85 & 130.72G & 14.13M & 13.71 \\
		w. Ours  & 5 & T & 78.91 & 89.19 & 136.01G &  14.16M & 13.37 \\ \noalign{\hrule height 1pt}
		STMENet & 5 & C & 73.40  & 85.96 & 41.92G & 9.85M &  11.95\\
		w. Ours  & 5 & C & 77.05 & 88.37 & 42.98G & 9.89M   & 11.87 \\
		\noalign{\hrule height 1pt}
	\end{tabular}}
\end{table}
To address the aforementioned challenges and prevent the introduction of new shortcut learning behaviors during model design, unlike almost all existing FPN designs, which rely on heuristic forward design, we start entirely from the perspective of backpropagation and propose \textbf{BP-FPN, a theoretically grounded macro-architecture} that comprises two modules: Gradient-Isolated Low-Level Shortcut (GILS) and Directional Gradient Regularization (DGR). The former introduces low-level semantic information to enhance small target localization without inducing shortcut learning, while the latter enforces the FPN to refine robust representations of small targets through gradient regularization, promoting consistency across fine-grained details from low-level semantics, contextual cues from high-level semantics, and intermediate semantic information. As shown in Fig. \ref{fig:Compared_With_Ex}, thanks to the proposed design, our method exceeds the baseline’s final performance using \textbf{only 32\% of its training cycles}. As illustrated in Fig. \ref{fig:Compared_With_mAP}, BP-FPN achieves significant performance improvements while introducing almost no additional overhead in terms of parameters, computation, or FPS. Moreover, as shown in Tab. \ref{Tab:General}, the proposed method can be seamlessly integrated with existing approaches as a plug-and-play module, delivering performance gains with nearly zero additional overhead.

We summarize the main contributions of the paper as follows:
\begin{itemize}
	\item We revisit moving infrared small target detection and reveal that its key bottleneck lies in the lack of robust intra-frame feature representations, which causes ambiguity rather than in spatio-temporal modeling itself.
	\item We propose BP-FPN, a backpropagation-driven feature pyramid architecture that achieves significant performance gains with negligible overhead. \textbf{To the best of our knowledge, it is the first FPN designed for this task entirely from the backpropagation perspective.}
	\item We design Gradient-Isolated Low-Level Shortcut (GILS) to effectively incorporate fine-grained details of small targets while avoiding shortcut learning and preserving generalization.  
	\item We introduce Directional Gradient Regularization (DGR) to enforce directional consistency among hierarchical features during backpropagation, and provide theoretical analyses demonstrating its effectiveness and generality.  
	\item Extensive experiments on multiple public datasets verify the effectiveness and plug-and-play nature of our method, consistently yielding notable improvements across various existing frameworks.  
\end{itemize}

\section{Related Work} \label{Section:Related_Work}
We briefly review related works from three aspects: moving infrared small target detection, shortcut learning and feature pyramid network.

\subsection{Moving Infrared Small Target Detection}
Infrared small targets inherently lack distinctive discriminative cues, and false alarms in cluttered backgrounds often exhibit intensity distributions nearly indistinguishable from those of true targets \cite{ying2025infrared}. As a result, incorporating temporal information has proven highly effective \cite{11130659}, making moving infrared small target detection a topic of growing research interest due to its superior performance in complex scenes \cite{chen2024sstnet}.

However, most existing approaches \cite{duan2024triple,chen2025motion,chen2025language} adopt the conventional paradigm of natural-scene video object detection \cite{10024313}, emphasizing spatio-temporal feature fusion \cite{9438625,9474502,9797768} and refinement \cite{10608186} while overlooking the fundamental disparity between moving infrared small target detection and general video object detection tasks. Consequently, even elaborate spatio-temporal learning schemes bring only limited performance gains \cite{tong2024st}.
Specifically, conventional video object detection methods typically \textbf{assume} that \cite{zhou2022transvod} some frames suffer from visual degradation, and that aggregating features from adjacent reference frames can enhance detection performance on key frames. However, this assumption \textbf{does not hold} for moving infrared small target detection, where the targets inherently lack explicit discriminative cues and share highly similar feature distributions with false-alarm sources \cite{liu2023infrared}. As a result, the targets may \textbf{remain in a degraded state across the entire video sequence}, and the intrinsic ambiguity of intra-frame features further propagates to inter-frame aggregation, leading to degraded or ambiguous spatio-temporal representations.

Motivated by the above observations, we move beyond the conventional spatio-temporal fusion paradigm and propose to strengthen intra-frame feature representations with minimal additional complexity. Remarkably, our approach introduces merely 32K learnable parameters yet yields significant performance improvements, which we further substantiate through theoretical analysis.
\subsection{Shortcut Learning}
Shortcut learning \cite{10250856} is a notorious problem in deep neural networks (DNNs) \cite{11127214}, where models tend to rely on superficial statistical regularities or spurious correlations within the dataset \cite{10106000} rather than capturing semantically meaningful structures \cite{geirhos2020shortcut}. Such reliance often leads to erroneous predictions \cite{shah2020pitfalls}, as the model fails to generalize beyond the dataset’s surface-level patterns \cite{hermann2020shapes}. In particular, this behavior is an inherent property of the model itself \cite{DBLP:conf/iclr/ScimecaOCPY22}, which emerges and is reinforced during the training process through gradient backpropagation \cite{DBLP:conf/iclr/HermannMFM24}.
Existing studies have mainly focused on characterizing the mechanisms of shortcut learning \cite{rahaman2019spectral, teney2024neural, mahapatra2022interpretability, wang2024manifoldron}, or mitigating its effects in classification tasks \cite{luo2021rectifying}, while \textbf{its impact on moving infrared small target detection remains largely unexplored}. This limitation is especially \textbf{critical}, as the task fundamentally depends on the model’s ability to generalize to unseen video sequences \cite{zhang2025beyond}. Given the intrinsically weak discriminative cues of small targets \cite{li2023direction}, successful detection relies on modeling inter-frame feature correlations to distinguish true targets from false alarms \cite{chen2024micpl}. However, small targets are often treated as low-level semantic cues that can be captured by shallow networks \cite{RDIAN}. As a result, shortcut learning tends to drive the model toward memorizing simplistic visual patterns present in the training data, thereby severely constraining its generalization capability.

To address the aforementioned challenge, our core insight is that since shortcut learning is closely related to gradient propagation in shallow layers \cite{8804390}, we design a regularization mechanism for the shallow-layer gradients to encourage the model to reduce its dependency on specific low-level visual patterns learned from the training data.

\subsection{Feature Pyramid Network}
The Feature Pyramid Network (FPN) \cite{lin2017feature} is a cornerstone in modern object detection frameworks, refining multi-scale backbone features and integrating contextual information across levels \cite{chen2025yolo}. While FPNs have proven effective in general object detection \cite{wang2024yoloh}, they encounter several challenges in moving infrared small target detection:
\begin{enumerate}[(1)]
	\item Balancing detection performance and computational efficiency. Complex architectures \cite{10681553,10146336,9261609} or high-resolution representations \cite{9947071} can improve small-target detection but are computationally expensive. Temporal feature aggregation further amplifies this cost \cite{10076469}, resulting in prohibitive overhead.
	\item Heuristic designs may not generalize. As illustrated in Fig. \ref{fig:Compared_With_Ex}, heuristic high-resolution representations can encourage shortcut learning, where models overfit shallow-layer visual patterns, ultimately degrading robustness.
	\item Consistent multi-level feature enhancement is difficult. Despite extensive research on multi-scale fusion \cite{10818495,10045738,11112613}, heuristic designs often fail to reliably strengthen target-specific features.
\end{enumerate}

To overcome these challenges, we propose BP-FPN, a backpropagation-driven feature pyramid network that departs from conventional heuristic designs. BP-FPN is theoretically grounded and globally optimized, enabling efficient enhancement of small-target features:
\begin{enumerate}[(1)]
	\item Exploiting extreme target sparsity. Leveraging compressive sensing theory \cite{baraniuk2010model}, learnable downsampling can preserve sparse target features, enhancing low-level semantic cues with minimal computational overhead. The proposed GILS module maintains primary shallow-layer gradients, mitigating shortcut learning.
	\item Gradient-driven multi-level feature enhancement. The DGR module enforces consistency across feature levels via backpropagated gradients, guiding the network toward improved small-target detection. We provide theoretical analyses for both top-down and bottom-up variants, validating the effectiveness of gradient-driven optimization.
\end{enumerate}
To our knowledge, BP-FPN is the \textbf{first} FPN designed for moving infrared small target detection entirely from a backpropagation perspective.

\section{Methodology}
\subsection{Overview}
The proposed method, YOLO-BP (\underline{\textbf{Y}}ou \underline{\textbf{O}}nly \underline{\textbf{L}}ook \underline{\textbf{O}}mni Gradient \underline{\textbf{B}}ack\underline{\textbf{P}}ropagation), is composed of baseline (SSTNet \cite{chen2024sstnet}) integrated with the proposed BP-FPN. BP-FPN is a macro-architecture derived entirely from the perspective of backpropagation, as shown in Fig. \ref{fig:Compared_With_Ex}.(d). Its core consists of two modules: the Gradient-Isolated Low-Level Shortcut and Directional Gradient Regularization. The former ensures the introduction of low-level detail information under low computational cost while stabilizing the gradient's dominant components. The latter guarantees the collaborative optimization of low-level, mid-level, and high-level semantic information, refining features that are suitable for small object detection. It is important to note that since BP-FPN is a macro-architecture that manifests in the model's topological structure, as shown in Tab. \ref{Tab:General}, it can seamlessly integrate with existing methods and share specific module implementations, providing a plug-and-play enhancement to the model's performance.
\begin{remark}[\textbf{Unified Notation}]
	Following common practice in modern hierarchical vision models (\eg, FPN-style architectures \cite{8099589}), we use an overloaded notation for simplicity, where $\mathbf{C}_i$ and $\mathbf{P}_i$ denote both stage-wise feature representations and their corresponding parameterized modules. The intended meaning is context-dependent and can be clearly inferred from the surrounding formulation.
\end{remark}
\begin{definition}[\textbf{Omni Gradient Backpropagation}]
	\label{Def:1}
	Omni Gradient Backpropagation denotes a globally coupled optimization regime, where gradients from all pyramid levels are jointly accumulated onto a shared set of parameters while shortcut gradients are excluded. Formally,
	\begin{equation}
		\frac{\partial\mathcal L}
		{\partial\boldsymbol{\theta}_{\rm share}}
		=
		\sum_{i=1}^{L}
		\frac{\partial\mathcal L}
		{\partial\mathbf P_i}
		\frac{\partial\mathbf P_i}
		{\partial\boldsymbol{\theta}_{\rm share}},
		\label{eq:omni}
	\end{equation}
	where $\boldsymbol{\theta}_{\rm share}$ denotes the shared parameters introduced by DGR. Meanwhile, GILS modifies the computation graph through
	\begin{equation}
		\widetilde{\mathbf C}_i
		=
		\operatorname{Detach}(\mathbf C_i),
		\qquad
		\frac{\partial\widetilde{\mathbf C}_i}
		{\partial\mathbf C_i}
		=
		0,
		\label{eq:detach}
	\end{equation}
	such that shortcut gradients are eliminated during backpropagation. Therefore, \emph{Omni} refers to globally coupled gradient propagation over all semantic pathways of the feature pyramid, rather than omnidirectional spatial filtering.
\end{definition}

\textbf{Training Mechanism of BP-FPN.}
Unlike conventional feature enhancement modules that explicitly manipulate forward activations, BP-FPN reshapes the optimization process by modifying the computational graph during training. Consequently, BP-FPN introduces no additional operations during inference, and the improved feature representation is entirely attributed to the optimized network parameters learned under the modified gradient propagation.

Let the feature map at the $l$-th pyramid level be denoted by $\mathbf{F}_l$. In a conventional FPN, the forward computation is
\begin{equation}
	\mathbf{F}_l
	=
	\phi_l
	(
	\mathbf{F}_{l-1},
	\mathbf{S}_l;
	\theta_l
	),
\end{equation}
where $\mathbf{S}_l$ denotes the low-level shortcut feature.
The corresponding parameter gradient is
\begin{equation}
	\frac{\partial\mathcal L}
	{\partial\theta}
	=
	\sum_{l=1}^{L}
	\frac{\partial\mathcal L}
	{\partial\mathbf F_l}
	\frac{\partial\mathbf F_l}
	{\partial\theta}.
	\label{eq:bp}
\end{equation}
Instead of directly manipulating feature activations, GILS modifies the computational graph by applying the stop-gradient operator to the shortcut branch,
\begin{equation}
	\widetilde{\mathbf S}_l
	=
	\operatorname{Detach}
	(
	\mathbf S_l
	),
\end{equation}
yielding
\begin{equation}
	\mathbf F_l
	=
	\phi_l
	(
	\mathbf F_{l-1},
	\widetilde{\mathbf S}_l;
	\theta_l
	).
	\label{eq:gils_forward}
\end{equation}
According to the chain rule,
\begin{equation}
	\frac{\partial\widetilde{\mathbf S}_l}
	{\partial\mathbf S_l}
	=
	0,
\end{equation}
which naturally blocks shortcut gradients during backpropagation while leaving the forward feature computation unchanged.

To further encourage hierarchy-consistent optimization, DGR shares the parameters among adjacent fusion blocks. Consequently, the gradient with respect to the shared parameters becomes
\begin{equation}
	\frac{\partial\mathcal L}
	{\partial\theta_s}
	=
	\sum_{l=1}^{L}
	\frac{\partial\mathcal L}
	{\partial\mathbf F_l}
	\frac{\partial\mathbf F_l}
	{\partial\theta_s},
	\label{eq:dgr}
\end{equation}
where $\theta_s$ denotes the shared parameters.
Therefore, BP-FPN modifies the parameter gradient as
\begin{equation}
	\left(
	\frac{\partial\mathcal L}
	{\partial\theta}
	\right)_{\rm BP}
	=
	\left\{
	\begin{aligned}
		&
		\sum_l
		\frac{\partial\mathcal L}
		{\partial\mathbf F_l}
		\frac{\partial\mathbf F_l}
		{\partial\theta_s},
		&&
		\theta=\theta_s,
		\\
		&
		\sum_l
		\frac{\partial\mathcal L}
		{\partial\mathbf F_l}
		\frac{\partial\mathbf F_l}
		{\partial\theta},
		&&
		\text{otherwise},
	\end{aligned}
	\right.
\end{equation}
which is subsequently optimized by the standard SGD update
\begin{equation}
	\theta_{t+1}
	=
	\theta_t
	-
	\eta
	\left(
	\frac{\partial\mathcal L}
	{\partial\theta}
	\right)_{\rm BP}.
\end{equation}
During inference, the network performs the same forward computation
\begin{equation}
	\mathbf F
	=
	f(\mathbf X;\theta^{*}),
\end{equation}
without requiring any gradient computation. Therefore, BP-FPN improves feature representation indirectly by learning a better parameter optimum through gradient reshaping during training, rather than introducing gradient feedback during inference.

\subsection{Gradient-Isolated Low-Level Shortcut}
\begin{figure}[!t]
	\centering
	\includegraphics[width=\linewidth]{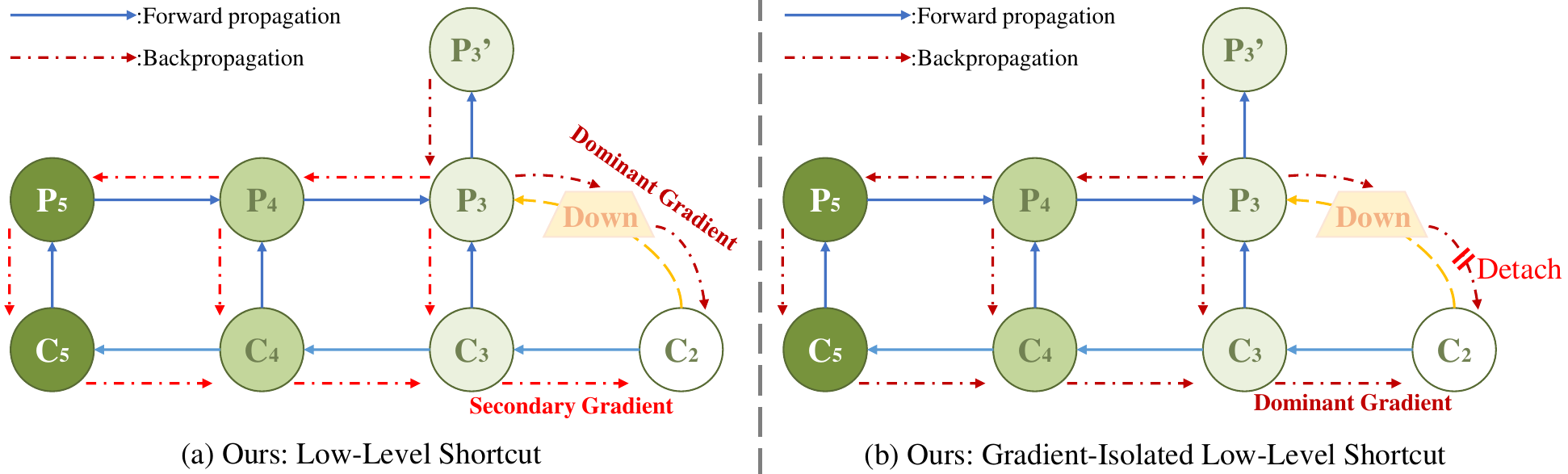}
	\caption{Comparison between the proposed Gradient-Isolated Low-Level Shortcut and the vanilla low-level shortcut. Through Gradient-Isolate, we ensure the consistency of the primary optimization direction. Moreover, since only the parameters of the downsampling module in the $C_2$ branch receive gradient updates at this stage, the downsampling operation is able to refine small-object details from high-resolution feature maps.}
	\label{fig:low_short}   
\end{figure}
\begin{table}[!t]
	\centering
	\caption{Impact of Gradient-Isolate on Low-Level Shortcut on The IRDST Dataset}
	\label{Tab:Low_Short}
	\setlength{\tabcolsep}{4.0pt}
	\begin{tabular}{c|c|cc|cc|c}
		\noalign{\hrule height 1pt}
		\textbf{Methods}                    & \textbf{Frames}                & $\textbf{mAP}_\textbf{50}$$\uparrow$                  & \textbf{F1}$\uparrow$                 & \textbf{Flops}$\downarrow$                 &\textbf{Params}$\downarrow$ & \textbf{PCR}$\uparrow$                   \\ \noalign{\hrule height 1pt}
		SSTNet & 5 & 71.55 & 85.11 & 123.59G & 11.95M & 0.578 \\
		w. Fig. \ref{fig:low_short}.(a)  & 5 & 81.40 & 90.65  & 128.69G &  11.98M & 0.632   \\ 
		w. Fig. \ref{fig:low_short}.(b)  & 5 & 82.86 & 91.48 & 128.69G &  11.98M & 0.643 \\ \noalign{\hrule height 1pt}
	\end{tabular}
\end{table}
The structure of the proposed Gradient-Isolated Low-Level Shortcut is shown in Fig. \ref{fig:low_short}.(b). The primary difference between this approach and the conventional method, which introduces target detail information from high-resolution feature maps by downsampling, is that we block the gradient backpropagation.

\begin{figure}[!t]
	\centering
	\includegraphics[width=\linewidth]{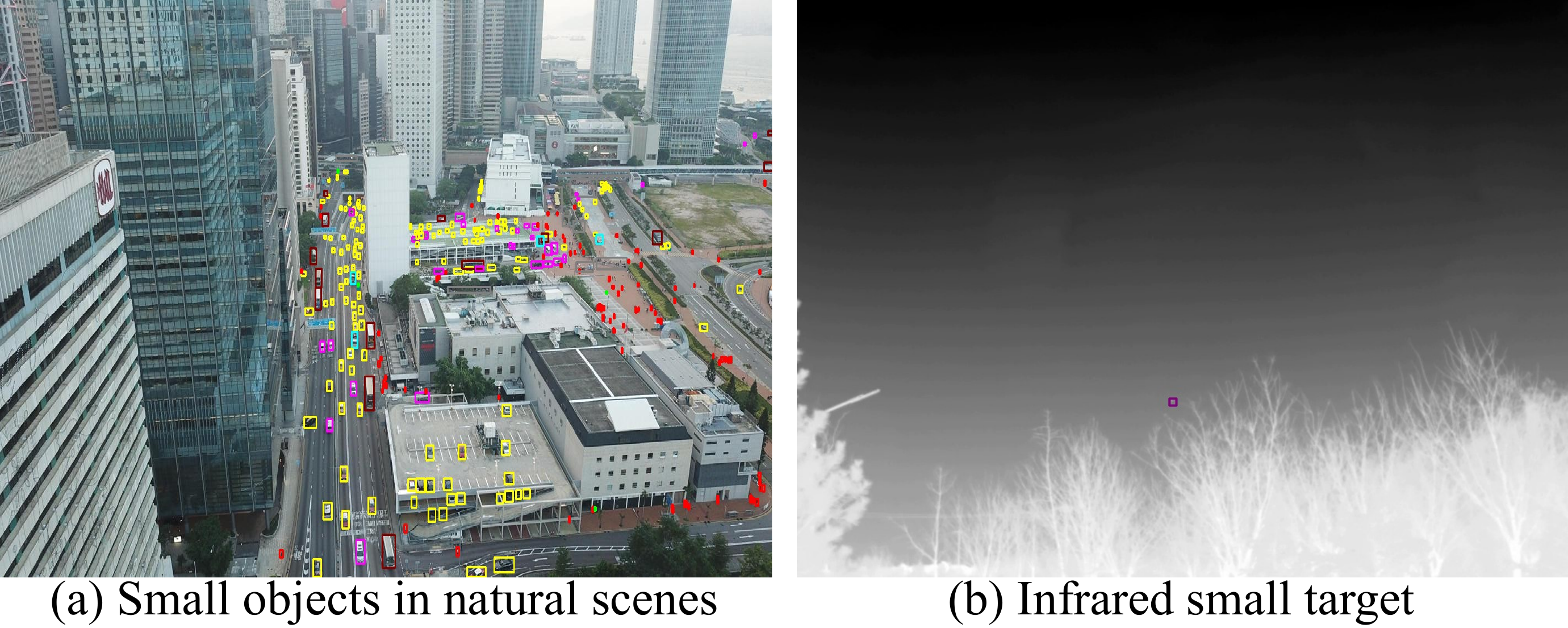}
	\caption{A comparison between small objects in natural scenes \cite{zhu2021detection} and infrared small targets \cite{RDIAN} shows that infrared targets exhibit extreme sparsity. According to compressive sensing theory \cite{baraniuk2010model}, under such conditions, suitable compression can preserve small-target information while simultaneously reducing the dimensionality of feature maps, obviating the need to adhere strictly to the Nyquist sampling principle.}
	\label{fig:small}   
\end{figure}

\noindent\textbf{Motivation.} The design motivation behind the Gradient-Isolated Low-Level Shortcut is twofold: (1) Low-level semantic information on high-resolution feature maps plays a crucial role in localizing small infrared targets \cite{10038696}. However, due to their extreme sparsity, directly processing them on high-resolution maps is computationally inefficient. The information bottleneck theory \cite{10637999} indicates that downsampling is an inherently information-compressive process. Under appropriate constraints, it can compel the network to learn how to discard background clutter while preserving target information \cite{11156113}. As shown in Fig. \ref{fig:small}, due to the extreme sparsity of the target itself, there exists a potential for retaining relevant information during the compression process; (2) As illustrated in Fig. \ref{fig:low_short}.(a), while directly introducing low-level shortcuts can enhance small target information, it alters the primary gradient flow on the $C_2$ feature map (similar to Fig. \ref{fig:Compared_With_Ex}.(b)  High-Resolution FPN), potentially leading the model to overfit specific visual patterns from the training set. This shortcut learning undermines generalization, making it essential to isolate gradient backpropagation to maintain the model's ability to generalize to unseen scenarios.

\noindent\textbf{Implementation Details.} Interestingly, from the perspective of backpropagation, introducing low-level semantic information without affecting the dominant gradient components and applying appropriate regularization to the downsampling module can be achieved simultaneously through a simple gradient blocking mechanism. Specifically, Given a high-resolution feature map $\mathbf{X}$, the processing flow is as follows:
\begin{equation}
	\mathbf{Y} = \text{Down}(\mathbf{X}.detach()),
\end{equation}
where $\mathbf{X}.detach()$ represents the feature map with gradient backpropagation blocked, preventing updates to the module that generates $\mathbf{X}$, and $\text{Down}(\cdot)$ is a simple downsampling module. Since $\mathbf{X}$ is a high-resolution feature map, directly applying convolutional operators on it is inefficient. Therefore, the specific implementation of the $\text{Down}(\cdot)$ module is as follows:
\begin{equation}
	\mathbf{X} = \text{PixelUnshuffle}_2(\mathbf{X}),
\end{equation}
\begin{equation}
	\mathbf{Y} = \text{Conv}_{1\times1}(\mathbf{X}).
\end{equation}
Here, $\text{PixelUnshuffle}_2(\cdot)$ is a function in PyTorch that reshuffles the feature map with a stride of 2 and concatenates it \cite{sunkara2022no}, transforming the original $\mathbf{X}\in\mathbb{R}^{C\times H\times W}$ into a lower-resolution representation $\mathbf{X}\in\mathbb{R}^{4C\times\frac{H}{2}\times\frac{W}{2}}$. At this point, the 1$\times$1 convolution serves to reduce the dimensionality, transforming $\mathbf{X}\in\mathbb{R}^{4C\times\frac{H}{2}\times\frac{W}{2}}$ into $\mathbf{Y}\in\mathbb{R}^{2C\times\frac{H}{2}\times\frac{W}{2}}$.

\noindent\textbf{Discussion.} In this section, we clarify the correspondence between implementation and motivation. First, since gradient backpropagation is blocked, the primary gradient source for the $C_2$ feature map remains unchanged, preventing shortcut formation (i.e., direct memorization of visual patterns). Meanwhile, updates to $C_2$ propagate to deeper layers (e.g., $C_3$), ensuring consistent optimization across the network. Thus, gradient blocking contributes to training stability.

Second, the information loss of feature maps is $\frac{2C \times \frac{H}{2} \times \frac{W}{2}}{C \times H \times H} = \frac{1}{2}$. However, due to spatial redundancy and inter-position correlation in images \cite{11156113}, the effective loss is substantially lower. Moreover, considering the extreme sparsity of infrared small targets, often modeled via $\ell_0(\cdot)$ constraints in model-driven methods \cite{Gao2013IPI}, the downsampling module preserves target information under appropriate constraints.
\begin{figure}[!t]
	\centering
	\includegraphics[width=\linewidth]{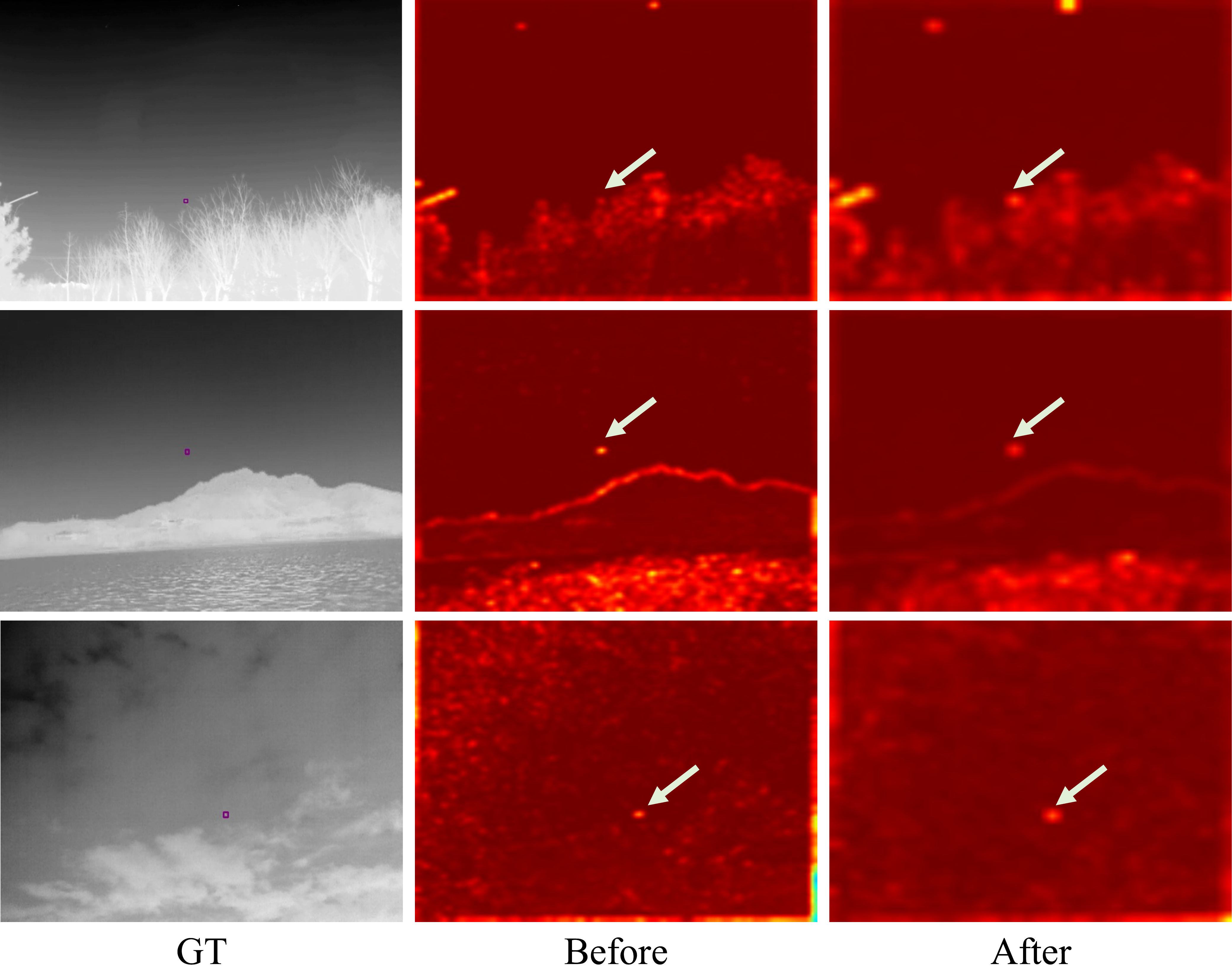}
	\caption{Visualization of feature maps before and after the GILS module. Owing to the introduced gradient detachment mechanism, the FPN is prevented from directly shortcutting the $C_2$ feature map for fine-grained detail extraction, thereby forcing the optimization process to rely on the hierarchical downsampling pathway. As a result, the learnable downsampling operation not only preserves small-target representations, but also effectively suppresses anomalous background responses, leading to more discriminative target-aware features.}
	\label{fig:GIShow}   
\end{figure}

Finally, as shown in Fig. \ref{fig:GIShow}, gradient blocking prevents the FPN from directly shortcutting $C_2$ for fine-grained details, forcing it to rely on the downsampling pathway. This implicitly regularizes the downsampling module, as its gradients are steered toward refining target details, thereby improving feature extraction for small object detection. As shown in Tab.~\ref{Tab:Low_Short}, Gradient-Isolate significantly improves performance.

\subsection{Directional Gradient Regularization}
\begin{figure}[!t]
	\centering
	\includegraphics[width=\linewidth]{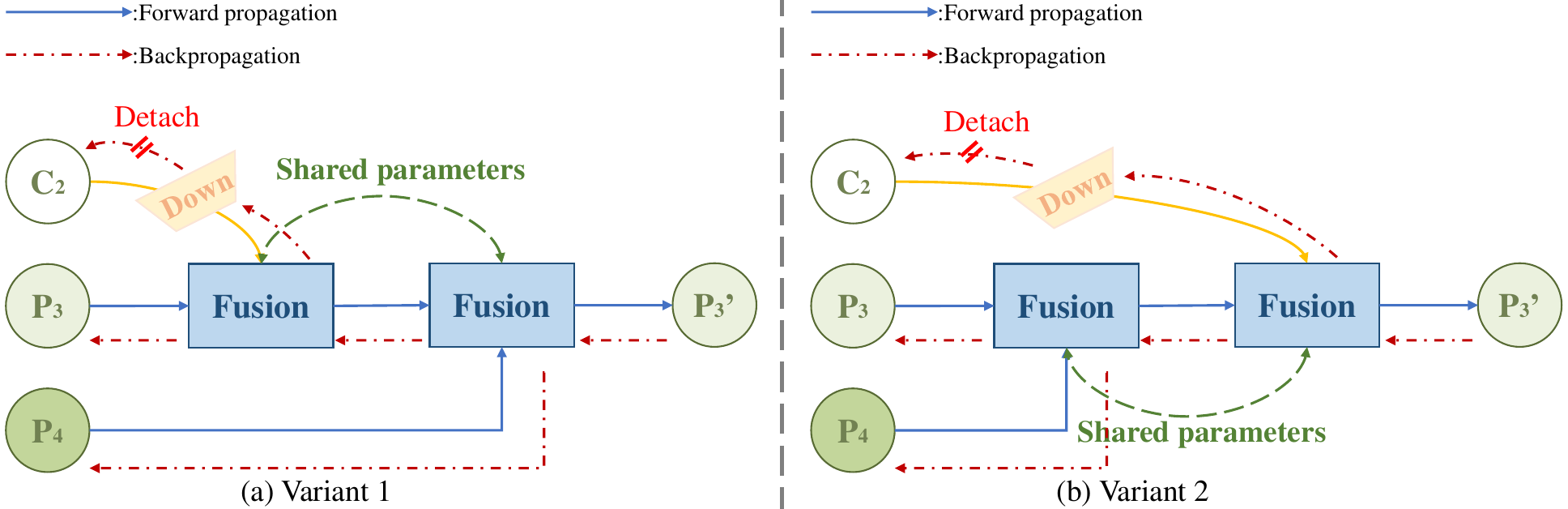}
	\caption{Variants of gradient regularization in different directions. Despite the high structural similarity between subgraph (a) and subgraph (b), the sequence of feature fusion results in gradient regularization along different directions, ultimately causing notable performance differences. \textbf{It should be noted that} the backpropagation directly yields the macro-level architecture, while the implementation of the Fusion module remains consistent with that of the baseline.}
	\label{fig:variants}   
\end{figure}
\begin{table}[!t]
	\centering
	\caption{Impact of Directional Gradient Regularization on Different Update Directions on The IRDST Dataset}
	\label{Tab:Direction}
	\setlength{\tabcolsep}{4.0pt}
	\begin{tabular}{c|c|cc|cc|c}
		\noalign{\hrule height 1pt}
		\textbf{Methods}                    & \textbf{Frames}                & $\textbf{mAP}_\textbf{50}$$\uparrow$                  & \textbf{F1}$\uparrow$                 & \textbf{Flops}$\downarrow$                 &\textbf{Params}$\downarrow$ & \textbf{PCR}$\uparrow$                   \\ \noalign{\hrule height 1pt}
		SSTNet & 5 & 71.55 & 85.11 & 123.59G & 11.95M & 0.578 \\
		w. Fig. \ref{fig:variants}.(a)  & 5 & 80.23 &  90.12 & 128.69G &  11.98M & 0.623   \\ 
		w. Fig. \ref{fig:variants}.(b)  & 5 & 82.86 & 91.48 & 128.69G &  11.98M & 0.643 \\ \noalign{\hrule height 1pt}
	\end{tabular}
\end{table}
The proposed Directional Gradient Regularization structure is illustrated in Fig. \ref{fig:variants}.(b). This structure represents a macroscopic topology that emerges naturally from the backpropagation process, serving to regularize the fusion among low-, mid-, and high-level semantic information. This regularization forces the fusion module to learn the portions of the features that are beneficial for small object detection across all three levels.

\noindent\textbf{Motivation.} Low-level semantics contribute to precise localization \cite{liu2023infrared}, high-level semantics provide contextual cues for discrimination \cite{zhang2025vision}, and mid-level semantics are known for their generalization ability and are often employed for cross-layer feature alignment \cite{tian2020pfenet}. Therefore, the fusion module must leverage the complementary advantages of these three levels to achieve robust small target detection. Nonetheless, despite numerous architectural refinements, FPN-based structures still face difficulties in realizing efficient and scalable feature fusion \cite{11181148,11114909,10749979}. Heuristic module designs tend to introduce excessive complexity while offering uncertain benefits \cite{9447743,10036005,9367228}, and their computational overhead is further amplified in video-based scenarios with dense temporal aggregation.

Our core insight is that instead of designing complex, difficult-to-apply modules, it is more effective to achieve implicit gradient regularization through parameter sharing. Specifically, when parameters are shared, from the perspective of backpropagation, the gradients of the fusion between high-level-mid-level and mid-level-low-level semantic information simultaneously influence the same module. This causes the final gradient of the module to be a superposition of these gradient components, strengthening the shared beneficial representations while suppressing inconsistent ones. This implicitly regularizes the fusion module, compelling it to focus on extracting the aspects from high-level, mid-level, and low-level information that are advantageous for small object detection.

For simplicity, consider the following setup: $\mathbf{X}_1$, 
$\mathbf{X}_2$, and $\mathbf{X}_3$ are three features to be fused. After processing, they all have the same shape. $\text{Fusion}(\cdot, \cdot;\boldsymbol{\theta})$ is the fusion module, where $\boldsymbol{\theta}$ represents its learnable parameters. The process of parameter-shared fusion operates as follows:
\begin{equation}
	\mathbf{T} = \text{Fusion}(\mathbf{X}_1, \mathbf{X}_2;\boldsymbol{\theta}_{share}),\label{Eq:e4}
\end{equation}
\begin{equation}
	\mathbf{O} = \text{Fusion}(\mathbf{X}_3, \mathbf{T};\boldsymbol{\theta}_{share}).\label{Eq:e5}
\end{equation}
Let the loss at the output $\mathbf{O}$ be denoted as $\mathcal{L}$ at this point, then the gradient of $\boldsymbol{\theta}$ with respect to the backpropagation is given by:
\begin{equation}
	\frac{\partial \mathcal{L}}{\partial \boldsymbol{\theta}} = \left(\frac{\partial \mathcal{L}}{\partial \mathbf{O}} \cdot \frac{\partial \mathbf{O}}{\partial \boldsymbol{\theta}} \right) + \left(\frac{\partial \mathcal{L}}{\partial \mathbf{T}} \cdot \frac{\partial \mathbf{T}}{\partial \boldsymbol{\theta}} \right). \label{Eq:Gra}
\end{equation}
Clearly, due to parameter sharing, the gradients in both fusion steps are updated with respect to the same learnable parameters 
$\boldsymbol{\theta}$, which forces the fusion module to simultaneously "attend to" the shared information across these three input features that contributes to reducing the loss.

We emphasize that the use of parameter sharing is not intended to maximize the expressive power of an individual fusion module. Instead, this design choice is motivated by the nature of the task: infrared small targets inherently lack discriminative information. Therefore, in moving infrared small target detection, the key lies in aggregating temporal features and exploiting their correlations to infer motion cues, rather than relying on the fused features to directly identify the target \cite{10034772}. Based on this insight, our objective is not to endow the fusion module with excessively strong feature recognition capability, but to establish a macroscopic gradient coupling loop that encourages the network to focus on cross-level consistency. Parameter sharing provides an effective means to realize such directional gradient regularization, thereby guiding the optimization process. The quantitative results in Tables \ref{Tab:Direction} and \ref{Tab:AB} further validate that this form of guidance offers both sufficient and effective optimization for the proposed framework.

\noindent\textbf{Implementation Details.} The fusion process is performed pairwise, which introduces a sequential relationship between the feature fusion steps. This results in two variants, as shown in Fig. \ref{fig:variants}. Although their structures are highly similar, the fusion order introduces a new form of regularization during backpropagation, leading to significant performance differences. We adopt the variant shown in Fig. \ref{fig:variants}.(b) for the actual implementation. Specifically, given the three feature maps $C_2$, $P_3$, and $P_4$ that we focus on, the feature fusion process is as follows:
\begin{equation}
	\mathbf{T} = \text{Fusion}(P_3, P_4; \boldsymbol{\theta}_{share}),
\end{equation}
\begin{equation}
	P_3' = \text{Fusion}(\mathbf{T}, \text{Down}(C_2.detach()); \boldsymbol{\theta}_{share}).
\end{equation}
Note that the macroscopic topology architecture is derived from backpropagation. Furthermore, since the baseline or other current methods for moving infrared small target detection also have corresponding feature fusion modules, and our implementation utilizes parameter sharing with cyclical reuse, we directly adopt the specific feature fusion implementation from the baseline.

\noindent\textbf{Discussion.} As shown in Tab. \ref{Tab:Direction}, the forward propagation processes of the two corresponding variants differ only in the order of feature fusion, yet there is a significant performance discrepancy between them. This section provides a theoretical analysis of this phenomenon. In fact, given the distinct physical semantics of the input features, changing the fusion order implicitly applies different forms of regularization during backpropagation. To investigate this, we further expand Eq. (\ref{Eq:Gra}) as follows:
\begin{equation}
	\frac{\partial \mathbf{O}}{\partial \boldsymbol{\theta}} = \frac{\partial \text{Fusion}_2}{\partial \mathbf{T}} + \frac{\partial \text{Fusion}_2}{\partial \mathbf{X}_3},
\end{equation}
\begin{equation}
	\frac{\partial \mathbf{T}}{\partial \boldsymbol{\theta}} = \frac{\partial \text{Fusion}_1}{\partial \mathbf{X}_1} + \frac{\partial \text{Fusion}_1}{\partial \mathbf{X}_2},
\end{equation}
\begin{align} \label{Eq:Loss}
	\frac{\partial\mathcal{L}}{\partial\boldsymbol{\theta}}& = \left( \frac{\partial\mathcal{L}}{\partial\mathbf{O}} \cdot \left( \frac{\partial \text{Fusion}_2}{\partial\mathbf{T}} \cdot \frac{\partial\mathbf{T}}{\partial\boldsymbol{\theta}} + \frac{\partial \text{Fusion}_2}{\partial\mathbf{X}_3} \cdot \frac{\partial\mathbf{X}_3}{\partial\boldsymbol{\theta}} \right) \right) \\ \nonumber
	& + \left( \frac{\partial\mathcal{L}}{\partial\mathbf{T}} \cdot \left(\frac{\partial \text{Fusion}_1}{\partial\mathbf{X}_1} \cdot \frac{\partial\mathbf{X}_1}{\partial\boldsymbol{\theta}} +  \frac{\partial \text{Fusion}_1}{\partial\mathbf{X}_2} \cdot \frac{\partial\mathbf{X}_2}{\partial\boldsymbol{\theta}} \right)\right), 
\end{align}
In this setup, $\text{Fusion}_1$ denotes the first fusion operation, as described in Eq. (\ref{Eq:e4}), while $\text{Fusion}_2$ represents the second fusion operation, as described in Eq. (\ref{Eq:e5}). From Eq. (\ref{Eq:Loss}), it is evident that $\mathbf{X}_3$ directly influences the output, whereas $\mathbf{X}_1$ and $\mathbf{X}_2$ have an indirect effect. Therefore, the impact of $\mathbf{X}_3$ on the parameters through $\boldsymbol{\theta}$ is more pronounced. At this point, by considering the specific implementations of the two variants shown in Fig. \ref{fig:variants}, we can derive the regularizations they impose as follows:
\begin{enumerate}[(1)]
	\item \textit{Variant1:} In Variant 1, $P_4$ plays a similar role as $\mathbf{X}_3$, which allows the fusion module to first discern the context in which the target is situated, thereby capturing the target’s characteristics. This behavior can be interpreted as the establishment of a scene-level prior.
	\item \textit{Variant2:} In Variant 2, the fusion module places greater emphasis on potential candidate target regions, and subsequently combines the contextual information surrounding these regions to form feature representations.
\end{enumerate}
\begin{figure}[!t]
	\centering
	\includegraphics[width=\linewidth]{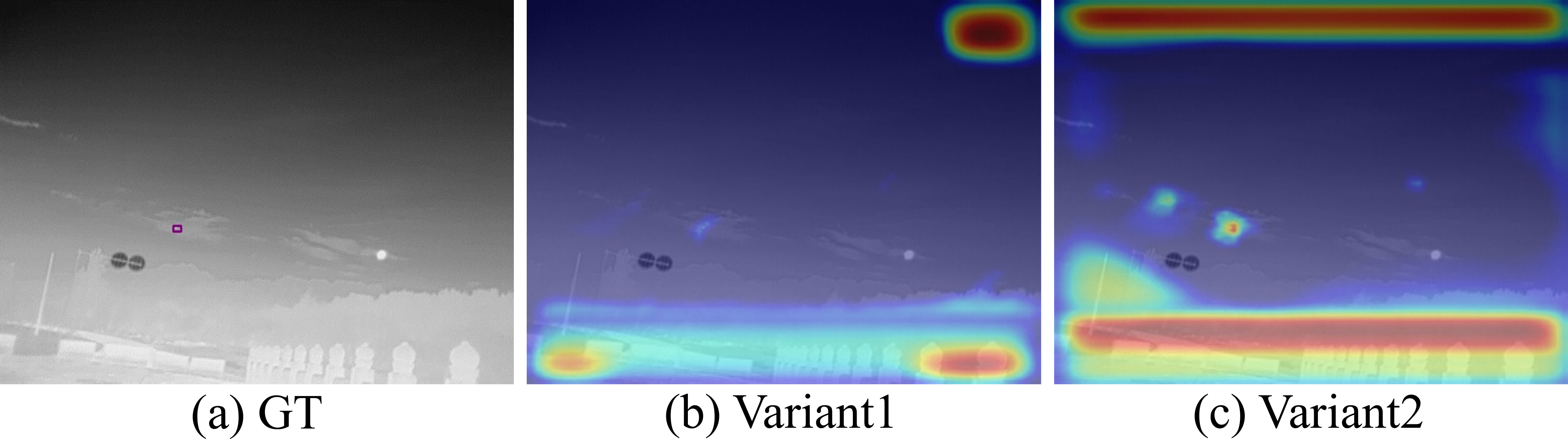}
	\caption{Grad-CAM \cite{8237336} comparison of different variants. The model distinguishes targets by comparing feature correspondences across different regions to capture variations in motion patterns \cite{chen2025language}, which also leads to attention on background areas \cite{chen2025motion}. Compared with Variant 1, Variant 2 maintains higher attention on potential target regions while still attending to the background, enabling the model to differentiate true targets from false alarms based on motion characteristics.}
	\label{fig:var}   
\end{figure}
As shown in Fig. \ref{fig:var}, Variant 2 attends to all regions that could potentially contain targets, enabling the model to differentiate true targets from false alarms by leveraging correspondences among temporal features. In contrast, Variant 1 focuses on certain structured background areas, paying relatively little attention to either false alarm sources or targets, which hampers target discrimination. This observation is consistent with the performance differences between the two variants obtained through backpropagation.

\section{Experiment}\label{Section:Experiment}
\subsection{Experimental Setup}
\subsubsection{Datasets} To comprehensively evaluate the performance of our method, we conduct experiments on three widely used public datasets for infrared small target detection in motion scenes: IRDST \cite{RDIAN}, ITSDT-15K \cite{duan2024triple}, and DAUB \cite{hui2019dataset}. These datasets cover a variety of scenarios and camera motion amplitudes. In many sequences, the targets are not only extremely small but also have low signal-to-clutter ratios (SCRs), and are often contaminated by numerous clutter sources that resemble the targets. Such characteristics make these datasets well suited for thoroughly assessing the robustness and effectiveness of detection algorithms.
\subsubsection{Evaluation Metrics} For performance evaluation, we adopt several widely used metrics in object detection, including Precision (Pr), Recall (Re), F1-score, $\text{mAP}_{50}$, and PCR (\eg, $\text{mAP}_{50}$ / GFlops) to assess the effectiveness of the proposed model.

\subsubsection{Implementation Details} In implementation, we followed the same settings as the baseline. The temporal window $T$ was uniformly set to 5, and the input image resolution was fixed at $512\times512$. We trained our YOLO-BP model for 100 epochs with a batch size of 4. The initial learning rate was set to 0.01, and stochastic gradient descent (SGD) was adopted as the optimizer, with a momentum of 0.937, a weight decay of $5\times10^{-4}$, and a learning rate decay factor of 0.1. During testing, only the predicted bounding boxes with confidence scores greater than 0.001 were retained. The intersection-over-union (IoU) threshold for non-maximum suppression (NMS) was set to 0.65. All experiments were conducted on a single NVIDIA V100 GPU.

\subsection{Comparison with State-of-the-Arts}
\begin{figure}
	\centering
	\includegraphics[width=0.95\linewidth]{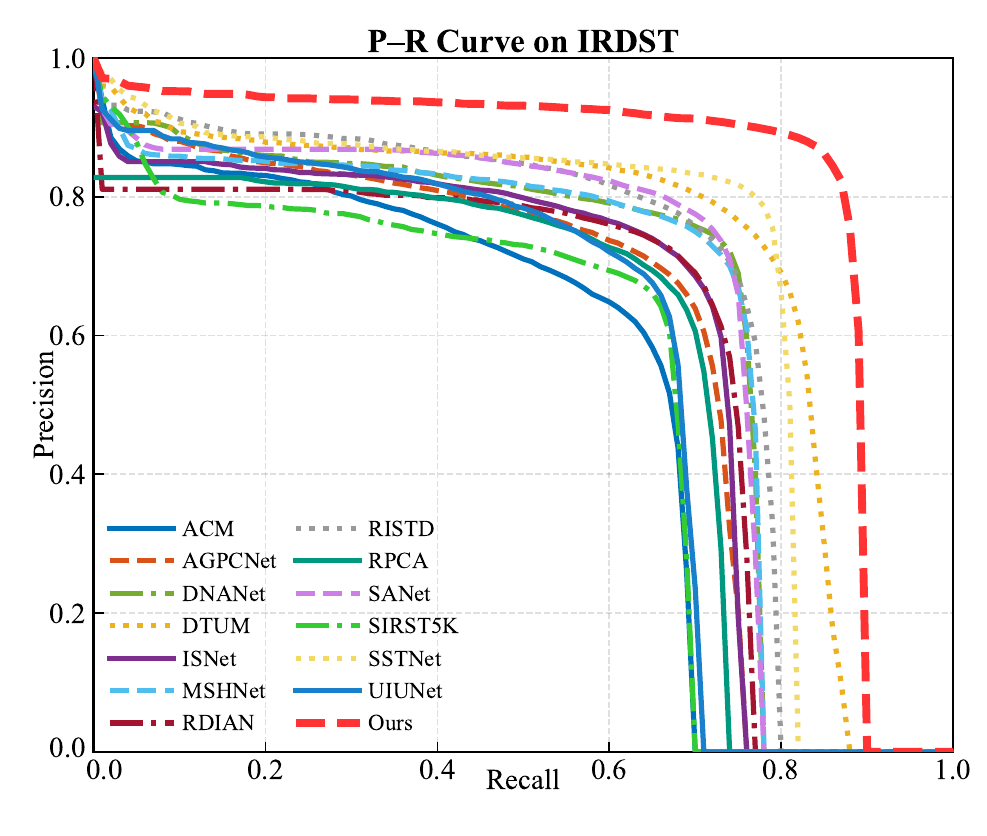}
	\caption{Precision–Recall Curve on IRDST.}
	\label{fig:PR}
\end{figure}
\begin{table*}[t!]
	\caption{Detection results achieved by different state-of-the-art methods. The best results are in \textbf{bold}, and the second-best results are \underline{underlined}. \textit{SF} and \textit{MF} refer to single-frame and multi-frame methods, respectively.}\label{tab:SOTA1}
	\centering
		\setlength{\tabcolsep}{6.0pt}
		\definecolor{mintgreen}{RGB}{204, 255, 204}   
		\definecolor{skyblue}{RGB}{204, 238, 255}     
		\definecolor{peachpink}{RGB}{255, 221, 238}   
		\begin{tabular}{cclcccccccccccc}
			\noalign{\hrule height 1pt}
			\multicolumn{3}{c}{\multirow{2}*{Methods}} & \multicolumn{4}{c}{\textbf{DAUB} \colorbox{mintgreen}{[Easy]}} & \multicolumn{4}{c}{\textbf{ITSDT-15K} \colorbox{skyblue}{[Medium]}} & \multicolumn{4}{c}{\textbf{IRDST} \colorbox{peachpink}{[Hard]}} \\
			\cline{4-15}
			\multicolumn{3}{c}{~}  &$\textbf{mAP}_\textbf{50}$     & \textbf{Pr} & \textbf{Re}  & \textbf{F1}& $\textbf{mAP}_\textbf{50}$ & \textbf{Pr} & \textbf{Re}& \textbf{F1} &$\textbf{mAP}_\textbf{50}$  & \textbf{Pr} & \textbf{Re}  & \textbf{F1}                  \\ \noalign{\hrule height 1pt}
			\multirow{6}*{\rotatebox{90}{Model-Driven}} & \multicolumn{1}{c}{\multirow{6}{*}{\rotatebox{90}{\textit{SF}}}}  
			& MaxMean \cite{deshpande1999max} \textit{(SPIE'99)} & 10.71 & 20.38 & 53.87 & 29.57 & 0.87 & 10.85 & 8.74 & 9.68 & 0.01 & 0.28 & 1.48 & 0.47 \\
			& \multicolumn{1}{c}{} & TopHat \cite{bai2010analysis} \textit{(PR'10)} & 16.99 & 21.69 & 79.83 & 34.11 & 11.61 & 27.21 & 43.07 & 33.35 & 1.81 & 18.22 & 10.60 & 13.40 \\
			& \multicolumn{1}{c}{}  & RLCM \cite{han2018infrared} \textit{(GRSL'18)} & 0.02 & 0.27 & 5.21 & 0.51 & 4.62 & 15.38 & 30.76 & 20.50 & 1.58 & 16.28 & 9.70 & 12.16 \\
			& \multicolumn{1}{c}{}  & HBMLCM \cite{8125583} \textit{(GRSL'19)} & 3.90 & 23.96 & 16.52 & 19.56 & 0.72 & 7.97 & 9.37 & 8.61 & 1.16 & 29.14 & 4.66 & 8.03\\  
			& \multicolumn{1}{c}{}  & RSTNN \cite{zhang2019infrared} \textit{(RS'19)} & 3.90 & 23.96 & 16.52 & 19.56 & 0.72 & 7.97 & 9.37 & 8.61 & 1.16 & 29.14 & 4.66 & 8.03\\
			& \multicolumn{1}{c}{}  & WSLCM  \cite{9130832} \textit{(GRSL'21)} & 1.37 & 11.88 & 11.57 & 11.73 & 2.36 & 16.78 & 14.53 & 15.58 & 1.69 & 20.87 & 8.70 & 12.28 \\  %
			\cline{2-15}
			\multirow{20}*{\rotatebox{90}{Data-Driven}}  & \multicolumn{1}{c}{\multirow{11}{*}{\rotatebox{90}{\textit{SF}}}}  & ACM \cite{Dai_2021_WACV} \textit{(WACV'21)} & 64.02 & 70.96 & 91.30 & 79.86 & 55.38 & 78.37 & 71.69 & 74.88 & 52.40 & 76.33 & 69.32 & 72.66 \\
			& \multicolumn{1}{c}{}  & RISTD \cite{hou2021ristdnet} \textit{(GRSL'22)} & 81.05 & 83.46 & 98.27 & 90.26 & 60.47 & 85.49 & 71.60 & 77.93 & 66.57 & 84.70 & 79.63 & 82.08 \\
			& \multicolumn{1}{c}{}  & ISNet \cite{ISNet} \textit{(CVPR'22)} & 83.43 & 89.36 & 94.99 & 92.09 & 62.29 & 83.46 & 75.32 & 79.18 & 59.78 & 80.24 & 75.08 & 77.58 \\
			& \multicolumn{1}{c}{}  & UIUNet \cite{UIUNet} \textit{(TIP'22)} & 86.41 & 94.46 & 92.03 & 93.23 & 65.15 & 84.07 & 78.39 & 81.13 & 56.38 & 80.95 & 70.29 & 75.25 \\
			& \multicolumn{1}{c}{}  & SANet \cite{zhu2023sanet} \textit{(ICASSP'23)} &87.12 & 93.44 & 94.93 & 94.18 & 62.17 & 87.78 & 71.23 & 78.64 & 64.54 & 84.29 & 77.02 & 80.49 \\
			& \multicolumn{1}{c}{}  & AGPCNet \cite{AGPCNet} \textit{(TAES'23)} & 76.72 & 82.29 & 94.43 & 87.95 & 67.27 & 91.19 & 74.77 & 82.16 & 59.21 & 79.47 & 75.51 & 77.44 \\
			& \multicolumn{1}{c}{}  & RDIAN \cite{RDIAN} \textit{(TGRS'23)} & 84.92 & 88.20 & 97.27 & 92.51 & 68.49 & 90.56 & 76.06 & 82.68 & 59.08 & 77.99 & 76.35 & 77.16\\
			& \multicolumn{1}{c}{}  & DNANet \cite{DNANet} \textit{(TIP'23)} & 89.93 & 92.49 & 98.27 & 95.29 & 70.46 & 88.55 & 80.73 & 84.46 & 63.61 & 82.92 & 77.48 & 80.11\\
			& \multicolumn{1}{c}{}  & SIRST5K \cite{lu2024sirst} \textit{(TGRS'24)} & 93.31 & 97.78 & 96.93 & 97.35 & 61.52 & 86.95 & 71.32 & 78.36 & 52.28 & 76.12 & 69.07 & 72.42 \\
			& \multicolumn{1}{c}{}  & MSHNet \cite{MSHNet} \textit{(CVPR'24)} & 85.97 & 93.13 & 93.12 & 93.13 & 60.82 & 89.69 & 68.44 & 77.64 & 63.21 & 82.31 & 77.64 & 79.91 \\
			& \multicolumn{1}{c}{}  & RPCANet \cite{RPCANet} \textit{(WACV'24)} & 85.98 & 89.38 & 97.56 & 93.29 & 62.28 & 81.46 & 77.10 & 79.22 & 56.50 & 77.77 & 73.80 & 75.73 
			\\ \cline{2-15} 
			& \multicolumn{1}{c}{\multirow{9}{*}{\rotatebox{90}{\textit{MF}}}}
			& DTUM \cite{li2023direction} \textit{(TNNLS'23)} & 85.86 & 87.54 & \textbf{99.79} & 93.26 & 67.97 & 77.95 & \underline{88.28} & 82.79 & 71.48 & 82.87 & 87.79 & 85.26 \\
			& \multicolumn{1}{c}{}  & TMP \cite{zhu2024tmp} \textit{(ESWA'24)} & 92.87 & 98.01 & 95.04 & 96.50 & 77.50 & 90.65 & 86.89 & 88.73 & 70.03 & 86.70 & 81.41 & 83.97 \\ 
			& \multicolumn{1}{c}{}  & ST-Trans \cite{10409231} \textit{(TGRS'24)} & 92.73 & 97.75 & 95.52 & 96.62 & 76.02 & 89.96 & 85.18 & 87.50 & 70.04 & 88.21 & 80.01 & 83.91\\
			& \multicolumn{1}{c}{}  & Tridos \cite{duan2024triple} \textit{(TGRS'24)} & \textbf{97.80} & \textbf{99.20} & \underline{99.67} & \textbf{99.43} & 76.72 & 91.81 & 84.63 & 88.07 & 73.72 & 84.49 & \underline{89.35} & \underline{86.85}\\
			& \multicolumn{1}{c}{}  & SSTNet \cite{chen2024sstnet} \textit{(TGRS'24)} & 95.59 & 98.08 & 98.10 & 98.09 & 76.96 & 91.05 & 85.29 & 88.07 & 71.55 & 88.56 & 81.92 & 85.11\\
			& \multicolumn{1}{c}{}  & STMENet \cite{peng2025moving} \textit{(EAAI'25)} & 92.04 & 97.20 & 95.70 & 96.45 & 77.33 & \underline{92.42} & 84.35 & 88.21 & 73.40 & 87.78 & 84.22 & 85.96 \\ 
			& \multicolumn{1}{c}{}  & MoPKL \cite{chen2025motion} \textit{(AAAI'25)} & - & - & - & - & \underline{79.78} & \textbf{93.29} & 86.80 & \underline{89.92} & \underline{74.54} & \underline{89.04} & 84.74 & 86.84\\ \rowcolor{yellow!30}
			& \multicolumn{1}{c}{}  & SSTNet w. Ours & \underline{96.43} & \underline{98.77} & 98.51 & \underline{98.64} & \textbf{82.20} & \underline{92.49} & \textbf{89.37} & \textbf{90.90} & \textbf{82.86} & \textbf{93.23} & \textbf{89.80} & \textbf{91.48}\\ \rowcolor{yellow!30}
			& \multicolumn{1}{c}{}  & $\Delta$ (Ours - SSTNet) $\uparrow$ & 0.84 & 0.69 & 0.41 & 0.55 & 5.24 & 1.44 & 4.08 & 2.83 & 11.31 & 4.67 & 7.88 & 6.37\\
			 \noalign{\hrule height 1pt}
		\end{tabular}
\end{table*}
\subsubsection{Quantitative Evaluation} The quantitative results are summarized in Tab. \ref{tab:SOTA1}. We primarily report the results as presented in the original papers of each method to ensure a consistent and fair comparison. Overall, the following conclusions can be drawn:
\begin{enumerate}[(1)]
	\item As the dataset difficulty increases, the performance gap between video object detection and single-frame object detection methods becomes progressively larger;
	\item Data-driven approaches significantly outperform model-driven ones;
	\item Existing techniques that focus on spatiotemporal feature fusion and refinement bring only marginal improvements;
	\item Our method achieves substantial performance gains by enhancing the feature representation of individual frames.
\end{enumerate}
Furthermore, the P–R curve shown in Fig. \ref{fig:PR} illustrates that our method achieves an excellent balance between precision and recall. In particular, compared with our baseline, it consistently delivers significant improvements in both precision and recall.

\begin{figure}[!t]
	\centering
	\includegraphics[width=\linewidth]{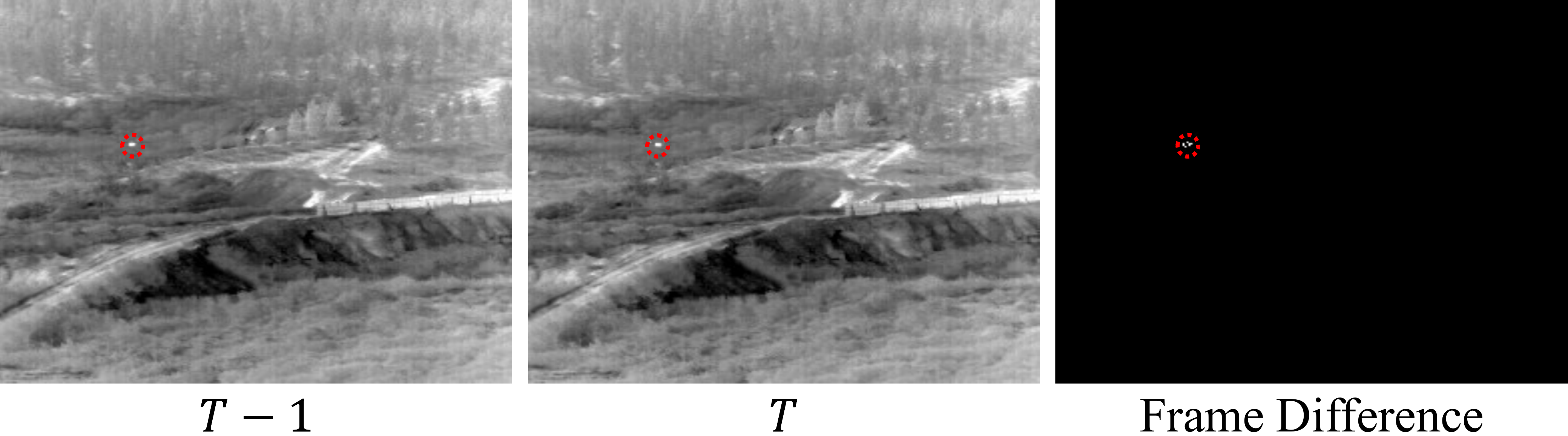}
	\caption{Representative samples from the DAUB dataset. Since the scenes are largely static and the target responses are relatively strong, simple frame differencing is often sufficient for target localization. Moreover, considering the inevitable annotation inaccuracies, the performance on this dataset is already close to saturation.}
	\label{fig:DAUB}   
\end{figure}
In addition, the quantitative results indicate that the performance improvement achieved by our BP-FPN is not uniform across the easy, medium, and hard datasets. Notably, the most pronounced gain is observed on the most challenging dataset. This can be attributed to the fact that, for the easy datasets (as shown in Fig. \ref{fig:DAUB}), object features are relatively clear and distinctive, and extensive feature refinement is less critical, as deep networks can already capture target characteristics effectively. Nevertheless, our method still yields noticeable improvements in this case. In contrast, for the hard datasets, the targets are faint and accompanied by numerous false alarms with similar characteristics. Under such challenging conditions, our approach provides more robust feature representations for the targets, leading to a significant improvement in detection performance.

\begin{table}[!t]
	\centering
	\caption{Complexity Comparisons of Inference on The IRDST Dataset}
	\label{tab:sota2}
	\setlength{\tabcolsep}{3.6pt}
	\begin{tabular}{l|c|cc|cc|c}
		\noalign{\hrule height 1pt}
		\textbf{Methods}                    & \textbf{Frames}                & $\textbf{mAP}_\textbf{50}$$\uparrow$                  & \textbf{F1}$\uparrow$                 & \textbf{Flops}$\downarrow$                 &\textbf{Params}$\downarrow$ & \textbf{PCR}$\uparrow$                   \\ \noalign{\hrule height 1pt}
		ACM \cite{Dai_2021_WACV} & 1 & 52.40 & 72.66 & 24.66G & 3.04M &2.124\\
		RISTD \cite{hou2021ristdnet} & 1 & 66.57 & 82.08 & 76.28G & 3.28M &0.872\\
		SANet \cite{zhu2023sanet} & 1 & 64.54 & 80.49 & 42.04G & 12.40M &1.535\\
		AGPCNet \cite{AGPCNet} & 1 & 59.21 & 77.44 & 366.15G & 14.88M &0.161\\
		ISNet \cite{ISNet} & 1 & 59.78 & 77.58 & 265.74G & 3.48M &0.224\\
		UIUNet \cite{UIUNet} & 1 & 56.38 & 75.25 & 456.70G & 53.06M &0.123\\
		RDIAN \cite{RDIAN} & 1 & 59.08 & 77.16 & 50.44G & 2.74M &1.171\\
		DNANet \cite{DNANet} & 1 & 63.61 & 80.11 & 135.24G & 7.22M &0.470\\
		SIRST5K \cite{lu2024sirst} & 1 & 52.28 & 72.42& 182.61G & 11.48M & 0.286 \\
		MSHNet \cite{MSHNet} & 1 & 63.21& 79.91 & 69.59G & 6.59M & 0.908\\
		RPCANet \cite{RPCANet} & 1 & 56.50 & 75.73 & 382.69G &3.21M & 0.147 \\
		DTUM \cite{li2023direction} & 5 & 71.48 & 85.26 & 128.16G & 9.64M & 0.557\\
		TMP \cite{zhu2024tmp} & 5 & 70.03 & 83.97 & 92.85G & 16.41M & 0.754\\
		STMENet \cite{peng2025moving} & 5 &73.40 & 85.96 & 41.92G & 9.85M &1.750\\
		ST-Trans \cite{10409231} & 5 & 70.04 & 83.91 & 145.16G & 38.13M & 0.482\\
		Tridos \cite{duan2024triple} & 5 & 73.72 & 86.85& 130.72G & 14.13M & 0.563\\
		MoPKL \cite{chen2025motion} & 5 & 74.54 & 86.84 & 119.64G & 9.46M & 0.623 \\ 
		SSTNet \cite{chen2024sstnet} & 5 & 71.55 & 85.11 & 123.59G & 11.95M & 0.578 \\ \rowcolor{yellow!30}
		Ours  & 5 & 82.86 & 91.48 & 128.69G &  11.98M & 0.643 \\ \noalign{\hrule height 1pt} \rowcolor{yellow!30}
		$\Delta$ $\uparrow$  & 5 & 11.31 & 6.37 & 5.1G &  0.03M & 2.217 \\
		 \noalign{\hrule height 1pt}
	\end{tabular}
\end{table}
Finally, we discuss the Complexity Comparisons of Inference, with the results summarized in Tab. \ref{tab:sota2}. As can be observed, integrating our proposed method leads to a significant improvement in model performance. Moreover, compared with the baseline, our approach achieves a higher performance–cost ratio (PCR), demonstrating that it can enhance performance in a more computationally economical manner. Unlike conventional approaches that improve accuracy by increasing model capacity, BP-FPN enhances optimization efficiency through gradient propagation, thereby achieving better detection performance without proportionally increasing model complexity.
\subsubsection{Qualitative Evaluation}
\begin{figure*}
	\centering
	\includegraphics[width=\linewidth]{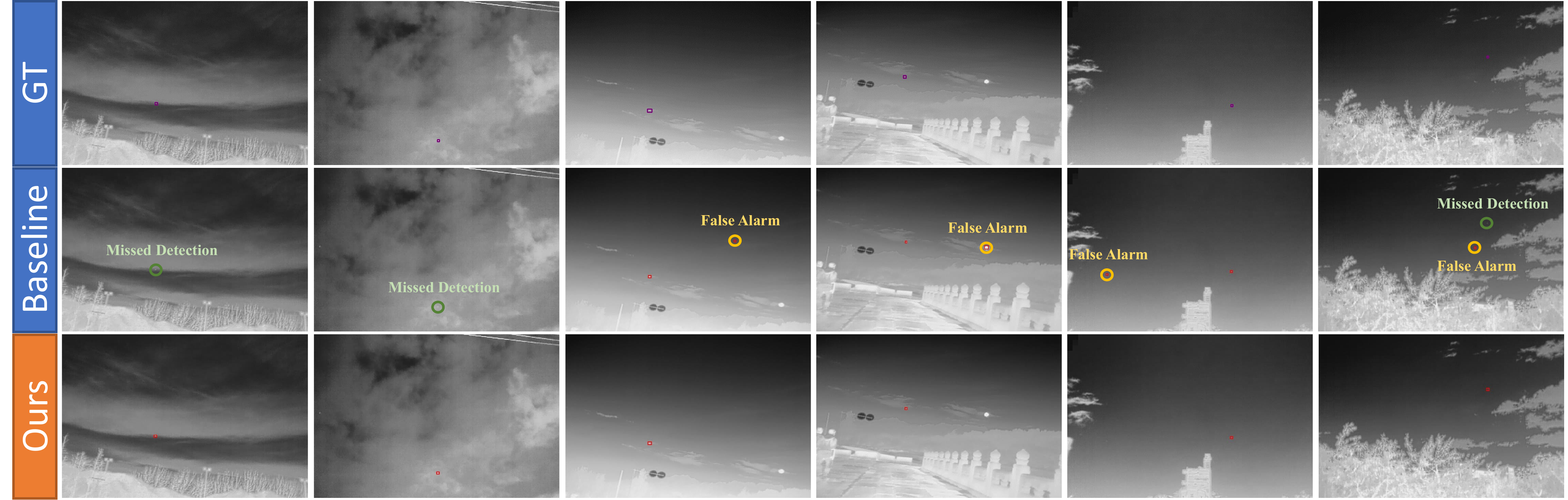}
	\caption{Visualization comparison of detection results between baseline and Ours across various challenging scenarios. [For better visual presentation, please zoom in the images.]}
	\label{fig:visual_more}   
\end{figure*}
Fig. \ref{fig:visual_more} presents qualitative comparisons between our method and the baseline under various challenging scenarios. It can be observed that our approach exhibits stronger robustness in complex environments. Owing to severe camera shake, many frames suffer from pronounced motion blur. Furthermore, the targets themselves are small and weak, lacking distinctive visual cues, which makes their appearance nearly indistinguishable from background clutter or false-alarm sources when viewed from a single-frame perspective. The baseline method, due to the absence of robust intra-frame feature representation, produces ambiguous features during temporal aggregation, ultimately leading to missed detections or false alarms.

\subsection{Ablation Study}
\begin{figure}[!t]
	\centering
	\includegraphics[width=\linewidth]{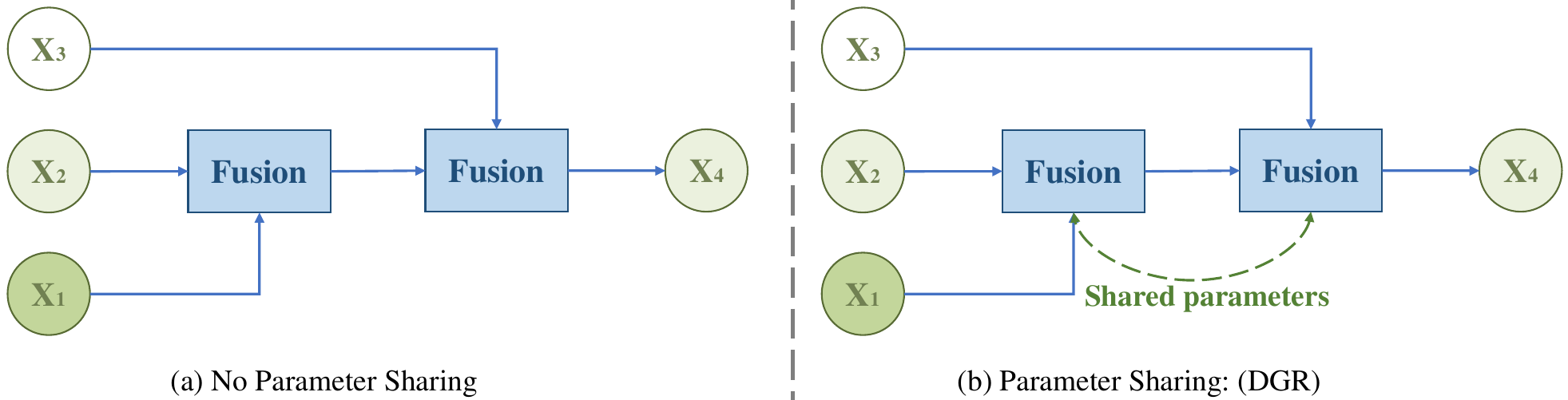}
	\caption{Comparison between the variant without using Directional Gradient Regularization (DGR) for feature fusion and DGR in the ablation study.}
	\label{fig:ABvariants}   
\end{figure}
\begin{table*}[!t]
	\centering
	\caption{The Ablation Experiments on The IRDST Dataset}
	\label{Tab:AB}
	\setlength{\tabcolsep}{4.7pt}
	\begin{tabular}{c|cc|c|cccc|ccc|cc}
		\noalign{\hrule height 1pt}
		\textbf{Strategy} &\textbf{GILS}   &   \textbf{DGR}               & \textbf{Frames}                & $\textbf{mAP}_\textbf{50}$ \textbf{(\%)} $\uparrow$     & \textbf{Pr} \textbf{(\%)} $\uparrow$ & \textbf{Re} \textbf{(\%)} $\uparrow$ & \textbf{F1}\textbf{(\%)} $\uparrow$                 & \textbf{Flops}$\downarrow$                 &\textbf{Params}$\downarrow$  & \textbf{FPS} $\uparrow$ & $\textbf{PCR}_\text{Params}$ $\uparrow$   & $\textbf{PCR}_\text{Flops}$ $\uparrow$                 \\ \noalign{\hrule height 1pt}
		(a) &\XSolidBrush & \XSolidBrush & 5 & 71.55 & 88.56 & 81.92 & 85.11 & 123.59G & 11.95M & 7.37 & 5.987 & 0.578 \\
		(b) &\Checkmark\kern-1.2ex\raisebox{1ex}{\rotatebox[origin=c]{125}{\textbf{--}}}\ddag & \Checkmark\kern-1.2ex\raisebox{1ex}{\rotatebox[origin=c]{125}{\textbf{--}}}\dag & 5 & 67.67 & 84.09 & 81.15 & 82.59 & 128.69G & 12.07M & 7.26 & 5.606 & 0.525 \\
		(c)&\CheckmarkBold & \Checkmark\kern-1.2ex\raisebox{1ex}{\rotatebox[origin=c]{125}{\textbf{--}}}\dag & 5 & 72.31 & 86.46 & 84.57 & 85.50 & 128.69G & 12.07M & 7.26 & 5.990 & 0.561 \\
		(d) &\Checkmark\kern-1.2ex\raisebox{1ex}{\rotatebox[origin=c]{125}{\textbf{--}}}\ddag & \CheckmarkBold & 5 & 81.40 & 91.81 & 89.52 & 90.65 & 128.69G & 11.98M & 7.26 & 6.794 & 0.632 \\
		(e)&\CheckmarkBold & \CheckmarkBold & 5 & 82.86 & 93.23 & 89.80 & 91.48 & 128.69G & 11.98M & 7.26 & 6.916 & 0.643 \\
		\noalign{\hrule height 1pt}
		\multicolumn{13}{l}{\footnotesize{(\dag) The feature fusion at this stage is performed using the approach illustrated in Fig. \ref{fig:ABvariants}.(a).}}\\
		\multicolumn{13}{l}{\footnotesize{(\ddag) At this stage, the incorporation of low-level semantic information is conducted following the scheme illustrated in Fig. \ref{fig:low_short}.(a).}}
	\end{tabular}
\end{table*}
In the ablation study, we evaluate the effectiveness of the proposed method on the highly challenging IRDST dataset by individually removing the Gradient-Isolated Low-Level Shortcut (GILS) and Directional Gradient Regularization (DGR) components. The results are summarized in Tab. \ref{Tab:AB}. \textbf{Please note that} our work does not involve designing new modules. Instead, we derive an FPN macro-architecture with theoretical guarantees through backpropagation, which spares us from laborious ablation studies. Moreover, variants of the proposed module have already been discussed in the Method section (The quantitative results are detailed in Tab. \ref{Tab:Low_Short} and \ref{Tab:Direction}.), and thus will not be repeated here.

\subsubsection{Effectiveness of the Gradient-Isolated Low-Level Shortcut Module} A comparison between Strategy (b) and Strategy (c) presented in Tab. \ref{Tab:AB} reveals a significant performance improvement with the introduction of the Gradient-Isolated Low-Level Shortcut Module. On the one hand, low-level semantic information contains critical cues for target localization. On the other hand, the gradient isolation mechanism preserves the dominant gradient direction of $C_2$ optimization, effectively preventing shortcut learning.

\begin{table}[!t]
	\centering
	\caption{Plug-and-Play Capability of Gradient-Isolated Low-Level Shortcut}
	\label{Tab:AB_Low}
	\setlength{\tabcolsep}{4.5pt}
	\begin{tabular}{c|c|cc|cc|c}
		\noalign{\hrule height 1pt}
		\textbf{Methods}                    & \textbf{Frames}                & $\textbf{mAP}_\textbf{50}$$\uparrow$                  & \textbf{F1}$\uparrow$                 & \textbf{Flops}$\downarrow$                 &\textbf{Params}$\downarrow$ & \textbf{PCR}$\uparrow$                   \\ \noalign{\hrule height 1pt}
		Fig. \ref{fig:Compared_With_Ex}.(c)  & 5 & 72.85 & 85.93 & 125.612G &  11.99M & 0.579 \\ 
		w. GILS & 5 & 74.51 & 86.81 &125.612G &  11.99M & 0.593
		\\\noalign{\hrule height 1pt}
	\end{tabular}
\end{table}
\subsubsection{Plug-and-Play Capability of Gradient-Isolated Low-Level Shortcut} Tab. \ref{Tab:AB_Low} demonstrates the plug-and-play capability of the Gradient-Isolated Low-Level Shortcut (GILS) module. By simply isolating the gradient flow in the downsampling branch of the Hourglass FPN \cite{wang2024yoloh}, without any additional cost, significant performance improvements are achieved. This not only validates the applicability of our GILS module but also highlights the detrimental effects of shortcut learning.

\begin{table}[!t]
	\centering
	\caption{Effectiveness of Learnable DownSampling in Gradient-Isolated Low-Level Shortcut}
	\label{Tab:AB_Low_Short_Down}
	\setlength{\tabcolsep}{4.0pt}
	\begin{tabular}{c|c|cc|cc|c}
		\noalign{\hrule height 1pt}
		\textbf{Methods}                    & \textbf{Frames}                & $\textbf{mAP}_\textbf{50}$$\uparrow$                  & \textbf{F1}$\uparrow$                 & \textbf{Flops}$\downarrow$                 &\textbf{Params}$\downarrow$ & \textbf{PCR}$\uparrow$                   \\ \noalign{\hrule height 1pt}
		MaxPooling & 5 & 77.81 & 88.66 & 127.69G & 11.95M & 0.609 \\
		AvgPooling  & 5 & 77.24 & 88.43  & 127.69G & 11.95M & 0.604   \\ 
		Ours  & 5 & 82.86 & 91.48 & 128.69G &  11.98M & 0.643 \\ \noalign{\hrule height 1pt}
	\end{tabular}
\end{table}
\subsubsection{Effectiveness of Learnable DownSampling in Gradient-Isolated Low-Level Shortcut}
Based on the analysis in the Methodology section, it is evident that the proposed DGR also has a regularizing effect on the downsampling module. Therefore, we discuss the effectiveness of learnable downsampling based on Strategy (e) shown in Tab. \ref{Tab:AB}. The ablation results are shown in Tab. \ref{Tab:AB_Low_Short_Down}, where it can be seen that learnable downsampling leads to a significant performance improvement. This is due to the gradient isolation, which forces the downsampling module to refine the target features during gradient propagation. Moreover, due to the extreme sparsity of small targets, their positional information can be effectively preserved by appropriate downsampling.

\subsubsection{Effectiveness of Proposed Directional Gradient Regularization}\label{Sec:DGR}
\begin{figure*}
	\centering
	\subfloat[Setting without gradient detachment and without the proposed DGR module (Strategy (b) in Tab. \ref{Tab:AB}).]{
		\includegraphics[width=\linewidth]{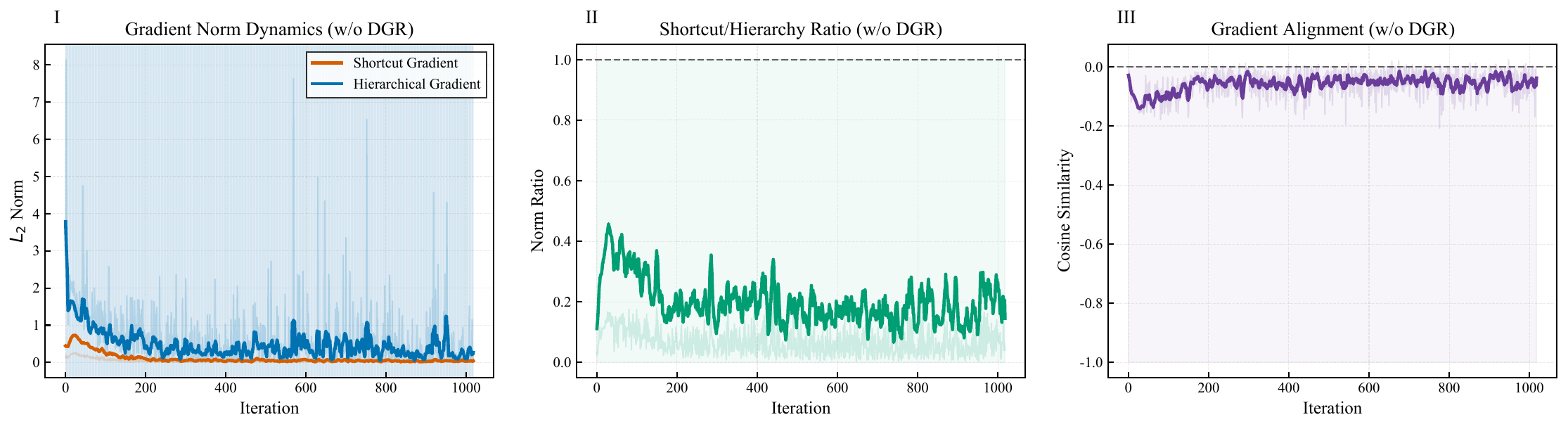}
		\label{fig:Shortcut}
	}
	\hfill
	\subfloat[Setting without gradient detachment but with the proposed DGR module (Strategy (d) in Tab. \ref{Tab:AB}).]{
		\includegraphics[width=\linewidth]{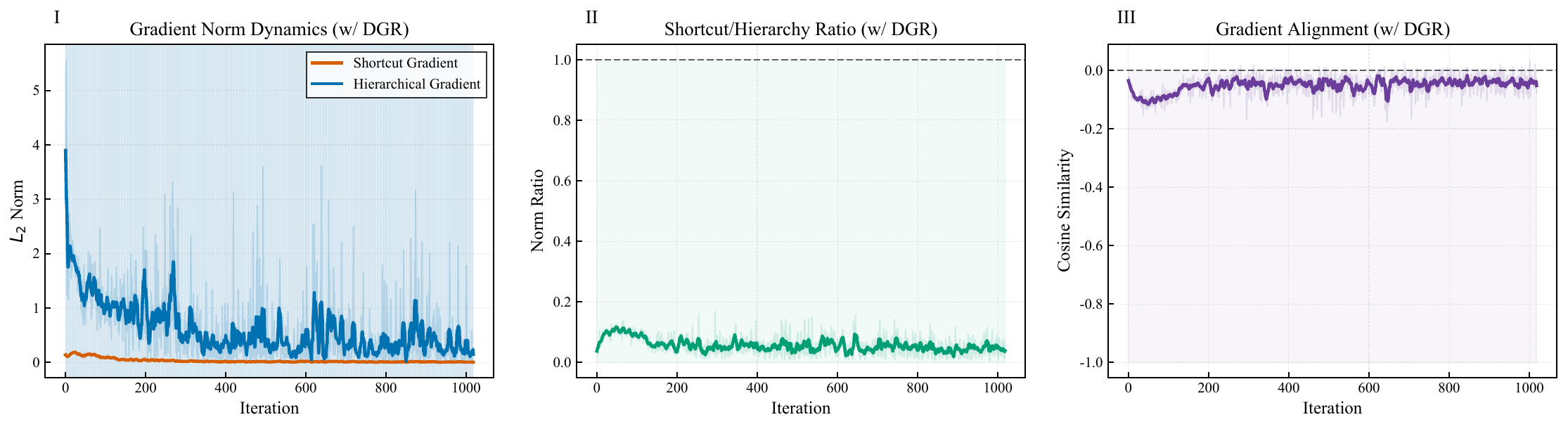}
		\label{fig:ShortcutwDGR}
	}
	\caption{Illustration of optimization dynamics (via hierarchical and shortcut gradients) in (a) without DGR and (b) with the proposed DGR module. Neither setting relies on hard gradient blocking. The cosine similarity between shortcut and hierarchical gradients remains consistently negative, indicating persistent antagonistic interactions between the two optimization pathways. Notably, DGR does not alter this directional conflict, as reflected by the unchanged cosine similarity. Instead, it suppresses only the magnitude of shortcut gradients, thereby reducing their relative influence during optimization. Consequently, although DGR reduces shortcut dominance, it remains insufficient to fully decouple the optimization dynamics between the two branches, further motivating the need for explicit gradient blocking in our framework. For better visualization, both the original gradient statistics and their smoothed trends are shown, where smoothing is performed using a moving-window average.}
	\label{fig:main}
\end{figure*}
Interestingly, when DGR is present, enabling the GILS gradient to backpropagate to $C_2$ leads to only a marginal performance drop, which contrasts sharply with the behavior observed in the High-resolution FPN shown in Fig. \ref{fig:Compared_With_Ex}. This can be attributed to the fact that, in addition to enforcing collaboration among features at different levels through gradient consistency, DGR also suppresses shortcut learning. To illustrate this point, we conduct a theoretical analysis of the variants shown in Fig. \ref{fig:ABvariants}. For convenience of exposition, the input variables are still denoted as $\mathbf{X}_1$, $\mathbf{X}_2$, and $\mathbf{X}_3$.

\textbf{No Parameter Sharing.} Without parameter sharing, the feature fusion processes in Eqs. (\ref{Eq:e4}) and (\ref{Eq:e5}) can be equivalently reformulated as:
\begin{equation}
	\mathbf{T} = \text{Fusion}(\mathbf{X}_1, \mathbf{X}_2; \boldsymbol{\theta}_1),
\end{equation}
\begin{equation}
	\mathbf{O} = \text{Fusion}(\mathbf{X}_3, \mathbf{T}; \boldsymbol{\theta}_2),
\end{equation}
where $\boldsymbol{\theta}_1 \neq \boldsymbol{\theta}_2$. At this point, the gradient backpropagation process for $\mathbf{X}_1$ is given by:
\begin{equation}
	\frac{\partial \mathcal{L}}{\partial \mathbf{X}_1} = \frac{\partial \mathcal{L}}{\partial \mathbf{O}} \cdot \frac{\partial \mathbf{O}}{\partial \mathbf{T}} \cdot
	\frac{\partial \mathbf{T}}{\partial \mathbf{X}_1}.
\end{equation}
In this case, the influence of $\mathbf{X}_3$ on $\mathbf{X}_1$ is manifested solely through this path in the chain rule, with no interaction terms arising from parameter sharing.

\textbf{Parameter Sharing.} Note that, under parameter sharing ($\boldsymbol{\theta}_1 \equiv \boldsymbol{\theta}_2$), the following relationship holds:
\begin{equation}
	\frac{\partial \mathcal{L}}{\partial \boldsymbol{\theta}_{share}} = \left(\frac{\partial \mathcal{L}}{\partial \mathbf{O}} \cdot \frac{\partial \mathbf{O}}{\partial \boldsymbol{\theta}_{share}} \right) + \left(\frac{\partial \mathcal{L}}{\partial \mathbf{T}} \cdot \frac{\partial \mathbf{T}}{\partial \boldsymbol{\theta}_{share}} \right),
\end{equation}
\begin{equation}
	\frac{\partial \mathcal{L}}{\partial \mathbf{T}} = \frac{\partial \mathcal{L}}{\partial \mathbf{O}} \cdot \frac{\partial \mathbf{O}}{\partial \mathbf{T}}.
\end{equation}
Therefore, we have:
\begin{align}
	&\frac{\partial\mathcal{L}}{\partial \mathbf{X}_1}=\underbrace{\frac{\partial\mathcal{L}}{\partial \mathbf{O}}\cdot \frac{\partial \text{Fusion}(\mathbf{X}_3, \mathbf{T}; \boldsymbol{\theta}_{share})}{\partial \mathbf{T}}\cdot\frac{\partial \mathbf{T}}{\partial \mathbf{X}_1}}_{\text{Direct path}}\\ \nonumber
	&+\underbrace{\frac{\partial\mathcal{L}}{\partial \mathbf{O}}\cdot\frac{\partial \text{Fusion}(\mathbf{X}_3,\mathbf{T};\boldsymbol{\theta}_{share})}{\partial\boldsymbol{\theta}_{share}}\cdot\frac{\partial \boldsymbol{\theta}_{share}}{\partial \mathbf{T}}\cdot\frac{\partial \mathbf{T}}{\partial \mathbf{X}_1}}_{\text{Indirect path with parameter sharing}}.
\end{align}
The analysis for $\mathbf{X}_2$ is analogous. It is noteworthy that $\mathbf{X}_1$, $\mathbf{X}_2$, and $\mathbf{X}_3$ are interdependent, with additional information channels established among them via the backbone network. As a result, $\mathbf{X}_1$ and $\mathbf{X}_2$ propagate gradients through the main backbone pathway to the update of $C_2$ associated with $\mathbf{X}_3$, effectively mitigating shortcut learning.
In summary, parameter sharing establishes a \textbf{\textit{cross-layer gradient coupling loop}} in the backpropagation process, compelling shallow features to respond to deep semantic error signals, thus alleviating the tendency toward shortcut learning.
\subsubsection{Comparison with Representative Lightweight Attention Modules} To further demonstrate the effectiveness of the proposed BP-FPN, particularly its ability to overcome the limitations of shortcut learning from a backpropagation perspective and enhance single-frame feature representations with nearly no additional computational overhead, we compare our method with conventional performance enhancement strategies based on lightweight attention mechanisms. Specifically, several representative lightweight attention modules are incorporated into the baseline for comparison.
\begin{table}[!t]
	\centering
	\caption{Comparison with Representative Lightweight Attention Modules}
	\label{Tab:AB_Attention}
	\setlength{\tabcolsep}{4.5pt}
	\resizebox{0.48\textwidth}{!}{\begin{tabular}{c|c|cc|cc|c}
		\noalign{\hrule height 1pt}
		\textbf{Methods}                    & \textbf{Publish}                & $\textbf{mAP}_\textbf{50}$$\uparrow$                  & \textbf{F1}$\uparrow$                 & \textbf{Flops}$\downarrow$                 &\textbf{Params}$\downarrow$ & \textbf{FPS}$\uparrow$                   \\ \noalign{\hrule height 1pt}
		CMO \cite{hou2024conv2former}  & TPAMI'24 & 68.51 & 83.18 & 129.82G & 12.54M & 7.17 \\
		LSK \cite{Li_2024_IJCV} & IJCV'24 & 63.59 & 79.95 & 126.85G & 12.25M & 6.69 \\
		SSR \cite{11069297}  & TCSVT'25 & 63.33 & 79.78 & 125.59G & 12.14M &  7.19 \\ 
		DHiF \cite{11373245} & TCSVT'26 & 66.66 & 82.08 & 123.64G & 11.95M &  5.53 \\ 
		Ours & - & 82.86 & 91.48 & 128.69G &  11.98M & 7.26
		\\\noalign{\hrule height 1pt}
	\end{tabular}}
\end{table}
As shown in Tab. \ref{Tab:AB_Attention}, simply inserting lightweight attention modules fails to bring performance gains and even results in a noticeable degradation. This may be attributed to the increased optimization difficulty introduced by attention mechanisms. As illustrated in Fig. \ref{fig:Compare}, in moving infrared small target detection, gradients flowing to the FPN originate from the total loss via spatiotemporally fused pathways, leading to a highly complex optimization process.

\subsubsection{Robustness Analysis Under Different Target Motion Speeds} To further verify that the proposed method enhances single-frame feature representations and consequently improves video target detection performance, we further evaluate the model under different target motion speeds (pixels per frame).
\begin{table}[!t]
	\centering
	\caption{Robustness Analysis Under Different Target Motion Speeds.}
	\label{Tab:VE}
	
	\resizebox{0.48\textwidth}{!}{
		\begin{tabular}{c c cc cc}
			\noalign{\hrule height 1pt}
			
			\multirow{2}{*}{No.} 
			& \multirow{2}{*}{Velocity} 
			& \multicolumn{2}{c}{Ours} 
			& \multicolumn{2}{c}{Baseline} \\
			
			\cmidrule(lr){3-4}
			\cmidrule(lr){5-6}
			
			&
			& $\textbf{mAP}_\textbf{50}$$\uparrow$
			& $\textbf{F1}(\%) \uparrow$
			& $\textbf{mAP}_\textbf{50}$$\uparrow$
			& $\textbf{F1}(\%) \uparrow$ \\
			
			\noalign{\hrule height 1pt}
			
			Seq. 22 & 4.560 & 100.0 & 100.0 & 98.08 & 99.60 \\
			Seq. 18 & 12.00 & 88.05 & 93.97 & 82.53 & 91.30 \\
			
			\noalign{\hrule height 1pt}\rowcolor[rgb]{0.9,0.9,0.9}
			Seq. 62 & 31.12 & 68.03 & 83.11 & 14.04 & 37.55 \\
			
			\bottomrule
		\end{tabular}
	}
\end{table}
As shown in Tab. \ref{Tab:VE}, the performance of the baseline degrades significantly with increasing target motion speed. In contrast, our method consistently achieves better results. This improvement can be attributed to the enhanced single-frame feature representation, which alleviates the difficulty of spatiotemporal correlation modeling in temporal scenarios.

Unlike motion-dependent approaches whose performance relies on accurate motion estimation, BP-FPN is inherently motion-agnostic. By suppressing shortcut-dominated gradients and reinforcing hierarchy-consistent semantic supervision during training, BP-FPN learns motion-invariant feature representations instead of velocity-specific patterns. Consequently, the optimization objective remains consistent across different target velocities, allowing the detector to maintain robust performance from nearly stationary targets to fast-moving targets.

\subsubsection{Synergistically Extended to Temporal Modeling}
To further demonstrate the extensibility of the proposed method, we generalize it to temporal modeling scenarios. As an FPN-based framework, temporal modeling requires the backbone to possess inherent temporal perception capability. Moreover, detector-oriented architectures typically rely on pretrained backbones to ensure stable optimization and satisfactory performance. However, in the infrared small target detection community, there still lacks publicly available backbone designs that simultaneously support temporal-aware representation learning and pretrained initialization.
To address this issue, we introduce an image-to-video adapter to transfer the image-based backbone to video settings. Specifically, we adopt the ST-Adapter \cite{NEURIPS2022_a92e9165}, which enables efficient spatiotemporal adaptation of pretrained 2D backbones via a lightweight residual formulation:
\begin{equation}
	\mathbf{F}^{t}_{l}
	=
	\mathbf{F}^{t}_{l,\mathrm{backbone}}
	+
	\mathcal{A}\!\left(\mathbf{F}^{t}_{l,\mathrm{backbone}}\right),
\end{equation}
where $\mathcal{A}(\cdot)$ is implemented using 3D convolutions to capture spatiotemporal dependencies. 
\begin{table}[!t]
	\centering
	\caption{Synergistically Extended to Temporal Modeling}
	\label{Tab:AB_3D}
	\setlength{\tabcolsep}{4.5pt}
	\resizebox{0.48\textwidth}{!}{\begin{tabular}{c|c|cc|cc|c}
			\noalign{\hrule height 1pt}
			\textbf{Methods}                    & \textbf{Publish}                & $\textbf{mAP}_\textbf{50}$$\uparrow$                  & \textbf{F1}$\uparrow$                 & \textbf{Flops}$\downarrow$                 &\textbf{Params}$\downarrow$ & \textbf{FPS}$\uparrow$                   \\ \noalign{\hrule height 1pt}
			ST-Adapter \cite{NEURIPS2022_a92e9165}  & NIPS'22 & 72.22 & 85.46 & 133.09G & 12.73M & 9.76 \\
			w. Ours & - & 75.63 & 87.87 & 138.21G &  12.75M & 9.70
			\\\noalign{\hrule height 1pt}
	\end{tabular}}
\end{table}
As shown in Tab. \ref{Tab:AB_3D}, our method consistently improves performance, further validating its extensibility to temporal modeling scenarios.

\subsubsection{Ablation analysis of information loss introduced by the GILS module}
As illustrated in Fig. \ref{fig:GIShow}, the representations after GILS processing may even exhibit enhanced target responses. This phenomenon arises because the compression process preserves task-relevant information while effectively suppressing irrelevant background interference. To quantitatively evaluate the information preservation capability before and after the GILS module, the signal-to-clutter ratio (SCR) and its corresponding gain (SCRG) are introduced \cite{Gao2013IPI}:
\begin{equation}
	\mathrm{SCRG} = \frac{\mathrm{SCR}_\mathrm{out}}{\mathrm{SCR}_\mathrm{in}}, \quad \mathrm{SCR} = \frac{\|\mu_t - \mu_b\|}{\sigma_b},
\end{equation}
where $\mu_t$ and $\mu_b$ denote the mean value of the target and background, respectively, $\sigma_b$ denotes the standard deviation of the background. Generally, the targets can be detected more easily in terms of infrared images with higher SCR. \textbf{\textit{Interestingly, after GILS processing, the SCRG can reach 1.37, exceeding 1}}. This indicates that GILS does not induce information loss; instead, it enhances target-relevant representations while suppressing background interference. This observation is consistent with the classical principle in representation learning that ``compression facilitates learning, and learning enables better compression'' \cite{11156113}, highlighting the beneficial role of GILS in refining task-discriminative features.

\subsubsection{Ablation Study on Backbone Generalization Across Different Architectures}
To further evaluate the generalization capability of the proposed method across different backbone architectures, we replace the CNN-based backbone in the baseline with Swin-Tiny \cite{9710580}.
\begin{table}[!t]
	\centering
	\caption{Ablation Study on Backbone Generalization Across Different Architectures}
	\label{Tab:Swin}
	\setlength{\tabcolsep}{4.5pt}
	\begin{tabular}{c|c|cc|cc|c}
		\noalign{\hrule height 1pt}
		\textbf{Methods}                    & \textbf{Frames}                & $\textbf{mAP}_\textbf{50}$$\uparrow$                  & \textbf{F1}$\uparrow$                 & \textbf{Flops}$\downarrow$                 &\textbf{Params}$\downarrow$ & \textbf{FPS}$\uparrow$                   \\ \noalign{\hrule height 1pt}
		Swin-Tiny  & 5 & 66.40 & 81.74 & 341.466G &  35.87M & 6.49 \\ 
		w. Ours & 5 & 73.40 & 87.80 &345.581G &  35.99M & 6.44
		\\\noalign{\hrule height 1pt}
	\end{tabular}
\end{table}
As shown in Tab. \ref{Tab:Swin}, the baseline performs noticeably worse with the Swin-Tiny backbone. This can be attributed to the limited inductive bias of Vision Transformers for modeling local target characteristics \cite{liu2023infrared}, together with the low-frequency preference induced by their large effective receptive fields, which weakens the high-frequency cues critical for infrared small target detection. Despite these limitations, integrating the proposed BP-FPN consistently yields significant performance gains, demonstrating its effectiveness and compatibility across heterogeneous backbone architectures.
\subsubsection{Discussion on Hard vs. Soft Gradient Detachment}
To demonstrate the necessity of the proposed Hard Gradient Detachment (HGD), we analyze the optimization dynamics between hierarchical (backbone-propagated) and shortcut gradients during training, as shown in Fig.~\ref{fig:main} and consistent with the analysis in Sec.~\ref{Sec:DGR}. As illustrated in Fig.~\ref{fig:Shortcut} (Strategy (b) in Tab.~\ref{Tab:AB}), the cosine similarity between the two gradient paths remains persistently negative, indicating conflicting optimization directions. This suggests that soft detach mechanisms are insufficient to resolve gradient conflicts.
Furthermore, as shown in Fig.~\ref{fig:ShortcutwDGR} (Strategy (d)), although the proposed DGR module reduces the magnitude of shortcut-induced interference, the antagonistic relationship between the two gradient paths still persists, motivating explicit Hard Gradient Detachment.
In addition, all hyperparameters are kept identical across methods to ensure fair comparison. Our method converges within 21 epochs, significantly faster than the baseline (65 epochs), indicating improved optimization stability.

\subsubsection{Impact of BP-FPN on Training Hyperparameters and Optimization Stability}
We further analyze how BP-FPN affects optimization dynamics during training. Based on the empirical observations in Fig.~\ref{fig:main}, the hierarchy-consistent gradients before and after gradient projection, denoted as $\mathbf{g}_h$ and $\mathbf{g}_h'$, satisfy
\begin{equation}
	\|\mathbf{g}_h'\|>\|\mathbf{g}_h\|,
	\label{eq:norm}
\end{equation}
while their directions remain highly aligned,
\begin{equation}
	\cos(\mathbf{g}_h',\mathbf{g}_h)\approx1.
	\label{eq:cos}
\end{equation}
The two observations jointly indicate that gradient projection primarily strengthens the hierarchy-consistent descent direction while introducing negligible directional deviation. Therefore, the projected gradient can be interpreted as
\begin{equation}
	\mathbf{g}_h'
	\approx
	\alpha\mathbf{g}_h,
	\qquad
	\alpha>1,
	\label{eq:projection}
\end{equation}
where the scaling factor $\alpha$ is directly implied by Eqs.~(\ref{eq:norm}) and (\ref{eq:cos}), rather than being introduced as an additional optimization hyperparameter.
Since all experiments are optimized using stochastic gradient descent (SGD), the parameter update follows
\begin{equation}
	\boldsymbol{\theta}_{t+1}
	=
	\boldsymbol{\theta}_{t}
	-
	\eta\mathbf{g},
	\label{eq:sgd}
\end{equation}
where $\eta$ denotes the learning rate. Substituting Eq.~(\ref{eq:projection}) into Eq.~(\ref{eq:sgd}) gives
\begin{equation}
	\boldsymbol{\theta}_{t+1}
	\approx
	\boldsymbol{\theta}_{t}
	-
	(\alpha\eta)\mathbf{g}_h.
	\label{eq:update}
\end{equation}
Equation~(\ref{eq:update}) indicates that BP-FPN effectively enlarges the optimization step only along hierarchy-consistent descent directions, rather than uniformly scaling all gradients. Therefore, faster convergence is achieved without sacrificing optimization stability.
The corresponding optimization behavior can be further characterized by the second-order Taylor expansion,
\begin{equation}
	\mathcal{L}(\boldsymbol{\theta}+\Delta\boldsymbol{\theta})
	\approx
	\mathcal{L}(\boldsymbol{\theta})
	+
	\mathbf{g}^{\top}\Delta\boldsymbol{\theta}
	+
	\frac{1}{2}
	\Delta\boldsymbol{\theta}^{\top}
	\mathbf{H}
	\Delta\boldsymbol{\theta},
\end{equation}
where $\mathbf{H}$ denotes the local Hessian matrix. Under SGD,
\begin{equation}
	\Delta\boldsymbol{\theta}
	=
	-\eta\mathbf{g},
\end{equation}
leading to
\begin{equation}
	\Delta\mathcal{L}
	\approx
	-
	\eta
	\|\mathbf{g}\|^{2}
	+
	\frac{\eta^{2}}{2}
	\mathbf{g}^{\top}
	\mathbf{H}
	\mathbf{g}.
	\label{eq:loss}
\end{equation}
Replacing $\mathbf{g}$ with the projected hierarchy-consistent gradient in Eq.~(\ref{eq:projection}) gives
\begin{equation}
	\Delta\mathcal{L}'
	\approx
	-
	\alpha^{2}\eta
	\|\mathbf{g}_h\|^{2}
	+
	\frac{\alpha^{2}\eta^{2}}{2}
	\mathbf{g}_h^{\top}
	\mathbf{H}
	\mathbf{g}_h.
	\label{eq:loss2}
\end{equation}
Compared with Eq.~(\ref{eq:loss}), Eq.~(\ref{eq:loss2}) indicates that BP-FPN increases the first-order descent while preserving nearly the same optimization direction, as verified by Eq.~(\ref{eq:cos}). Consequently, the optimizer achieves a larger loss reduction per iteration without introducing significant oscillatory updates. Therefore, BP-FPN improves optimization efficiency by reshaping the gradient field presented to SGD rather than modifying the optimizer itself, enabling faster and more stable convergence throughout training.

\subsection{Limitations and Future Work}
\begin{figure}
	\centering
	\includegraphics[width=\linewidth]{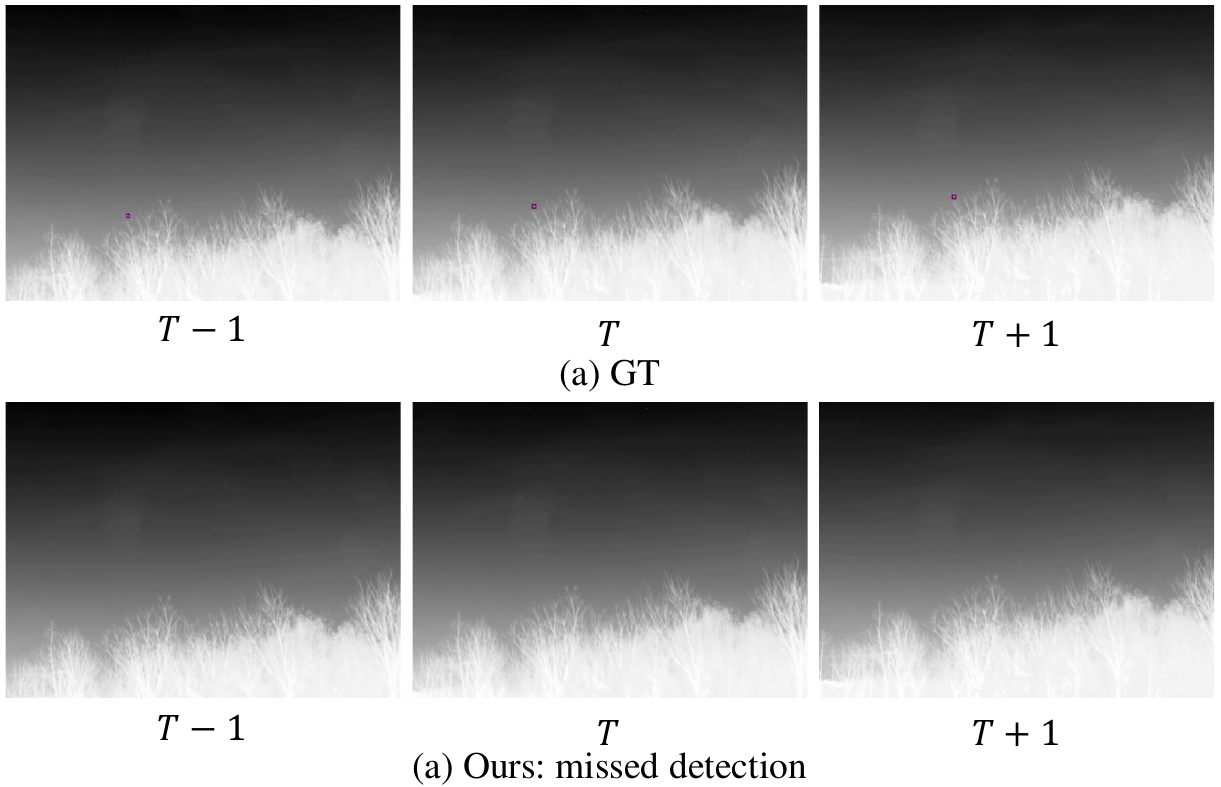}
	\caption{Typical failure cases. The target is small and dim, with a surrounding environment highly similar to background false-alarm sources. Due to its very slow and localized motion, it is mistakenly suppressed as a false alarm, leading to a missed detection. [For better visual presentation, please zoom in the images.]}
	\label{fig:Failed}
\end{figure}
Typical failure case is illustrated in Fig. \ref{fig:Failed}. Although the proposed method enhances the robustness of intra-frame features and thus improves the performance of moving infrared small target detection, the task is inherently a video-based detection problem, in which inter-frame correlation modeling remains crucial. In this case, the target itself lacks discriminative information and moves only within a small region adjacent to a false-alarm source. Consequently, the temporal feature aggregation module misclassifies it as a false alarm, leading to a missed detection. Future work may consider integrating explicit temporal correlation modeling \cite{10876405} or motion-aware priors \cite{7060702} to further mitigate such failures.

\begin{figure}
	\centering
	\includegraphics[width=\linewidth]{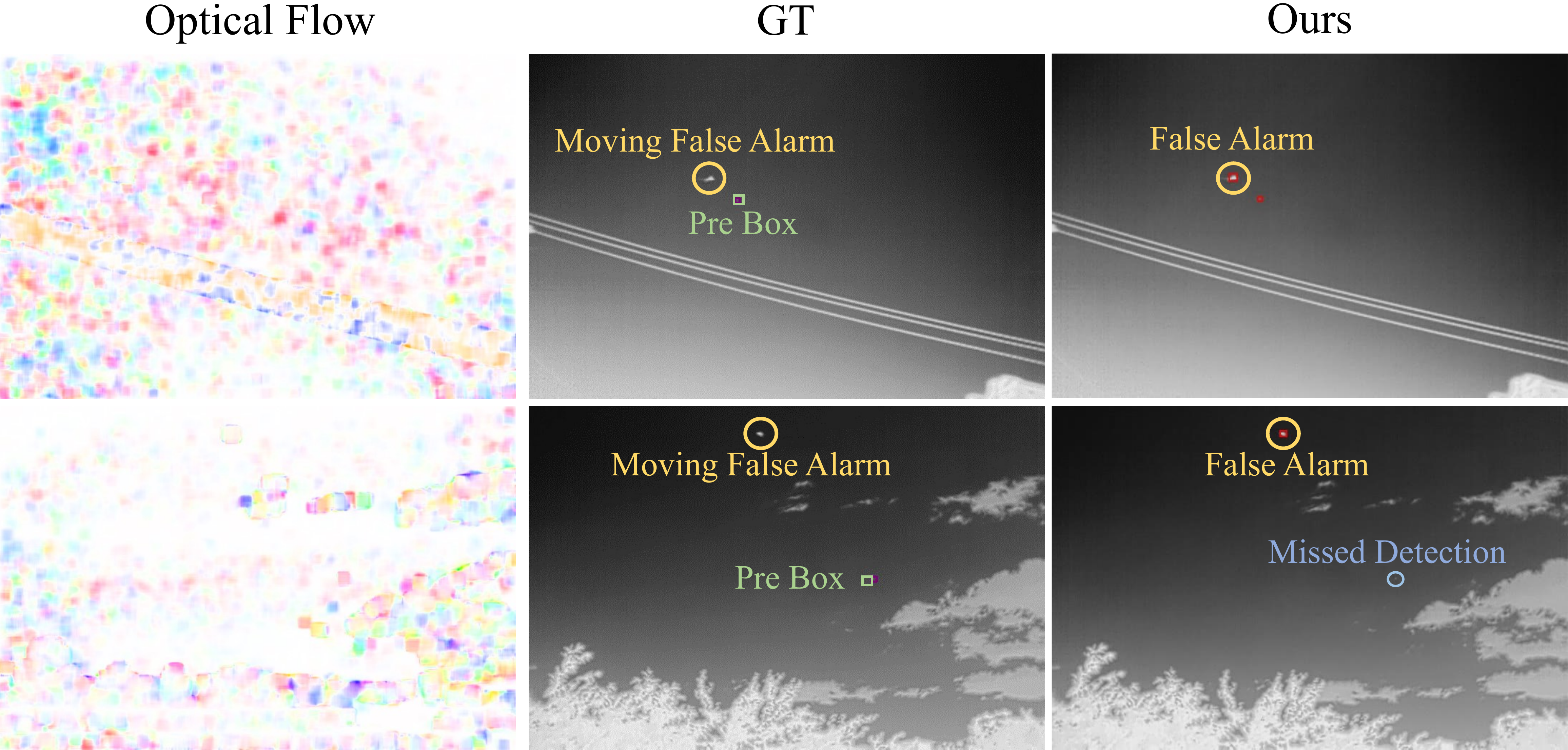}
	\caption{Typical failure cases. Moving false‑alarm sources in the background. [For better visual presentation, please zoom in the images.]}
	\label{fig:Failed1}
\end{figure}
Another typical failure arises from moving false‑alarm sources in the background. The proposed method fails here because it only enhances single‑frame features, whereas spatio‑temporal modeling still relies on the baseline. Recent work shows that when target, platform, and background all undergo complex motion, implicit spatio‑temporal aggregation is prone to being dominated by background information \cite{zhang2026decoupled}. Thus, more powerful spatio‑temporal modeling is needed in future work.

Finally, recent work \cite{zhang2026erf} shows that the ordering of effective receptive fields significantly affects infrared small target detection, and a theoretical scheduling framework has been introduced. Nevertheless, this framework only handles single‑frame cases; how to generalize it to video sequences is a worthwhile avenue for future research.

\section{Conclusion}\label{Section:Conclusion}
In this work, we revisited moving infrared small target detection and revealed that the primary bottleneck lies in the lack of robust per-frame feature representations rather than in spatio-temporal modeling. To address this, we proposed BP-FPN, a backpropagation-driven feature pyramid architecture, which integrates GILS for incorporating fine-grained low-level details without inducing shortcut learning, and DGR for enforcing hierarchical feature consistency during backpropagation. This principled design enables significant performance improvements with negligible computational overhead.
Extensive experiments on multiple public datasets demonstrate that BP-FPN can be seamlessly integrated as a plug-and-play module into existing frameworks, consistently establishing new state-of-the-art results. Our study highlights the importance of theoretically grounded feature representation design for small-object detection in complex infrared video scenarios, providing both practical solutions and insights for future research in this area.

\bibliographystyle{IEEEtran}

\begin{IEEEbiography}  [{\includegraphics[width=0.9in,height=1.125in,clip]{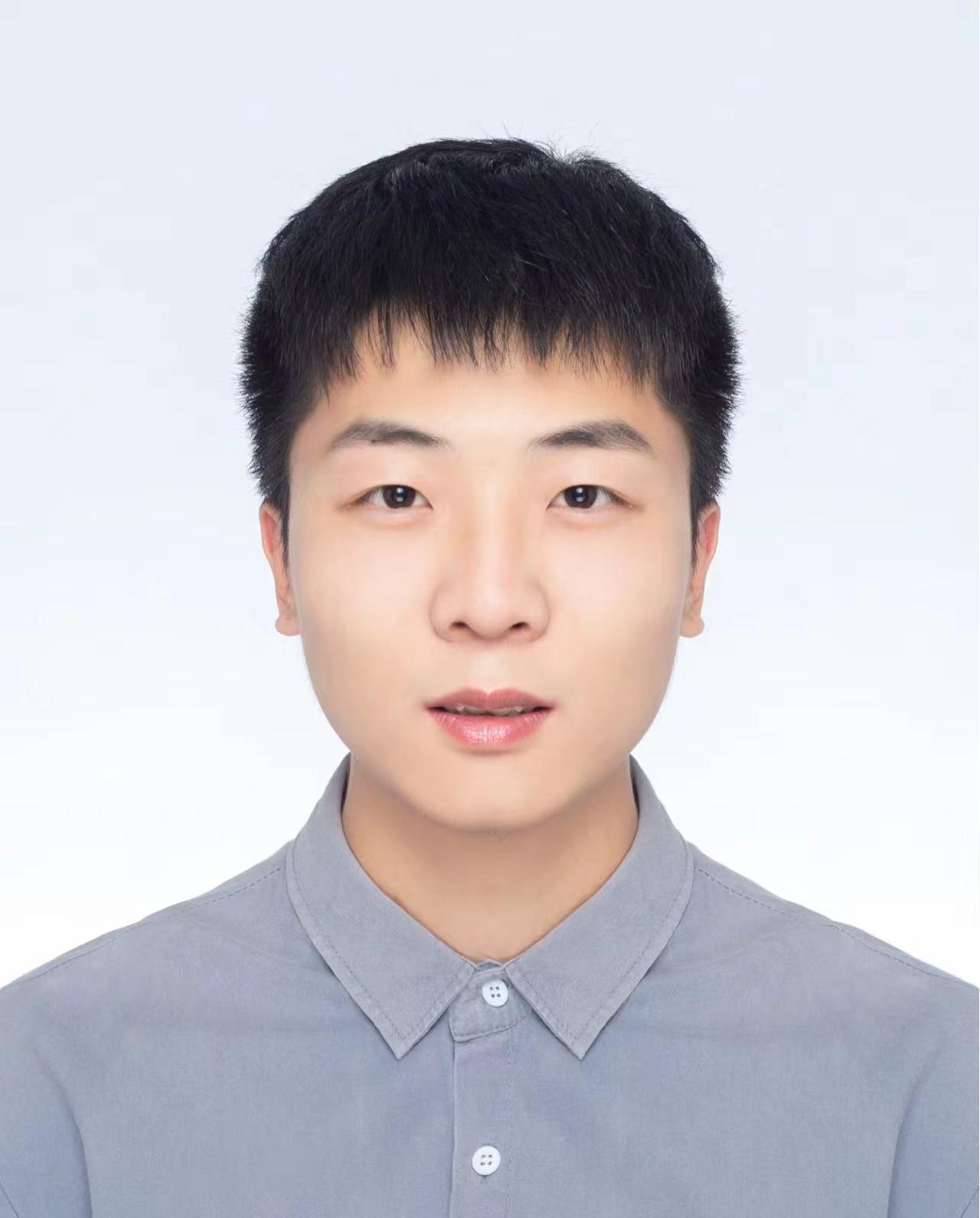}}] 
	{Guoyi Zhang} received the B.E. degree from Hefei University of Technology, Hefei, China, in 2023. He is currently pursuing the M.E. degree with the School of Aeronautics and Astronautics, Sun Yat-sen University, Shenzhen, China.\\
	His current research interests include representation learning and infrared small target detection. 
\end{IEEEbiography}

\begin{IEEEbiography}  [{\includegraphics[width=0.9in,height=1.125in,clip,keepaspectratio]{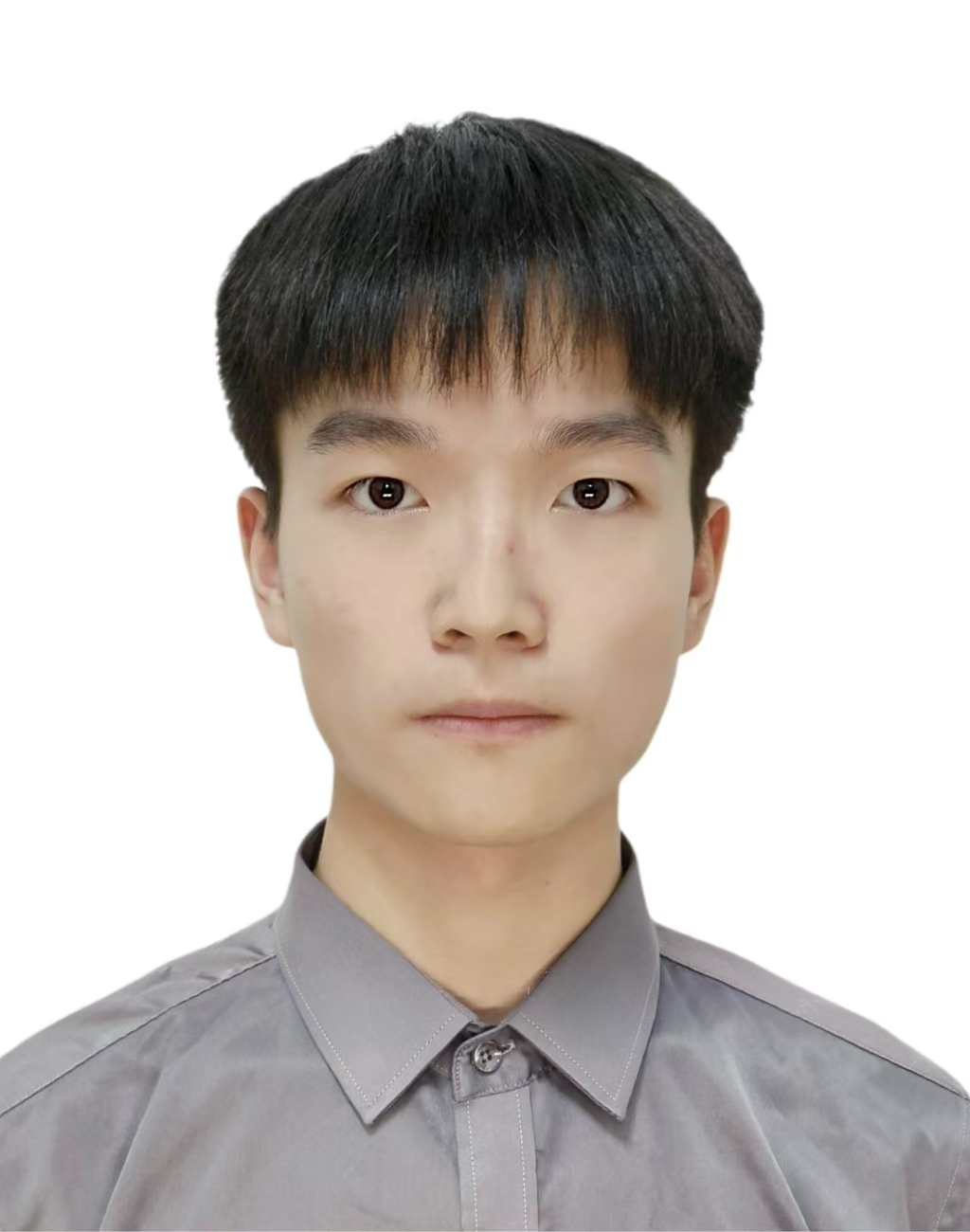}}] 
	{Guangsheng Xu} received the B.E. degree from Sun Yat-sen University, Shenzhen, China, in 2023. He is currently pursuing the M.E. degree with the School of Aeronautics and Astronautics, Sun Yat-sen University, Shenzhen, China.\\
	His current research interests include human pose estimation and human shape reconstruction.
\end{IEEEbiography}

\begin{IEEEbiography}  [{\includegraphics[width=1.0in,height=1.25in,clip,keepaspectratio]{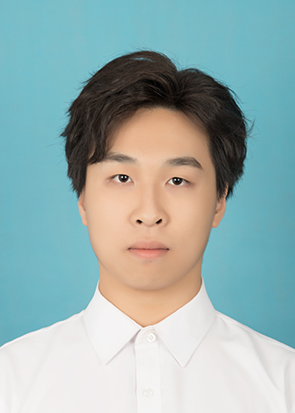}}] 
	{Siyang Chen} received the B.E. degree from Sun Yat-sen University, Shenzhen, China, in 2023. He is currently pursuing the M.E. degree with the School of Aeronautics and Astronautics, Sun Yat-sen University, Shenzhen, China.\\
	His research interests include space resident object detection and infrared small target detection.
\end{IEEEbiography}

\begin{IEEEbiography}  [{\includegraphics[width=1in,height=1.25in,clip,keepaspectratio]{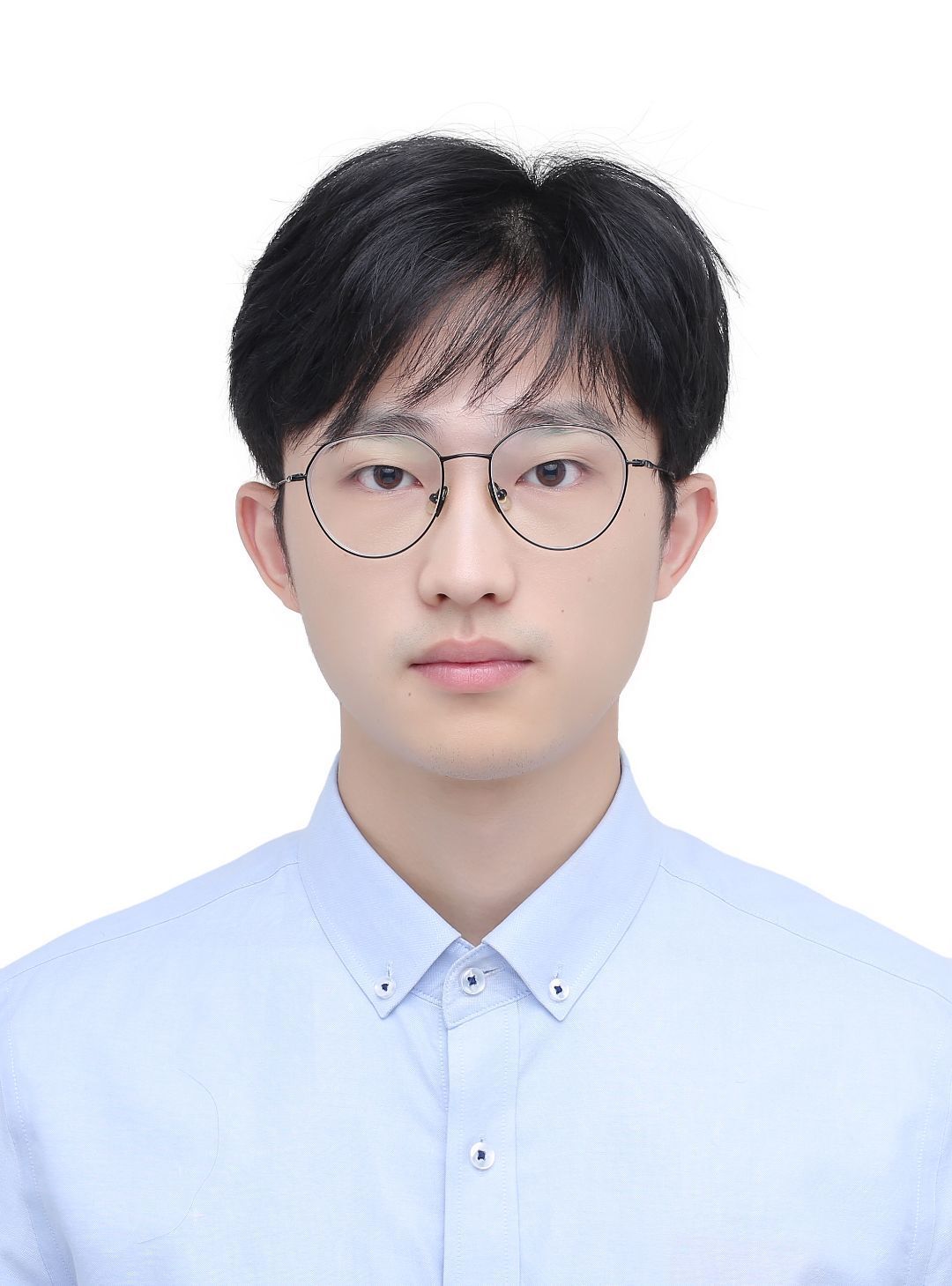}}] 
	{Han Wang} eceived the B.E. degree in mechanical engineering from Nanjing Agricultural University, Nanjing, China, in 2020, and the M.E. degree in mechanical engineering from China Agriculture University, Beijing, China, in 2022. He is currently working toward the Ph.D. degree with the School of Aeronautics and Astronautics, Sun Yat-sen University, Shenzhen, China.\\
	His research interests include astronomical image processing and space debris detection.
\end{IEEEbiography} 

\begin{IEEEbiography}  [{\includegraphics[width=0.9in,height=1.125in,clip,keepaspectratio]{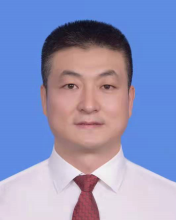}}] 
	{Xiaohu Zhang} received the Ph.D. degree in aeronautical and astronautical science and technology from National University of Defense Technology, Changsha, China, in 2006.\\
	He is currently a Professor with the School of Aeronautics and Astronautics, Sun Yat-sen University, Shenzhen, China. His research interests include aircraft visual perception, computer vision, and photogrammetry. 
\end{IEEEbiography} 
\newpage

\vfill

\end{document}